\documentclass[11 pt]{article}
\usepackage{graphicx}
\usepackage{amsmath}
\usepackage{mathrsfs}
\usepackage{hyperref}
\usepackage{amsthm}
\usepackage[toc,page]{appendix}

\def\R{\mathbf{R}}
\def\N{\mathbf{N}}
\def\(#1){[\hbox{$\mkern1mu\thickmuskip=\thinmuskip#1\mkern1mu$}]} 
\def\ast{\mathop{\hbox{\lower 1.5pt\hbox{$\buildrel x\over *$}}}}

\DeclareMathOperator*{\argmin}{arg\,min}

\newtheorem{theorem}{Theorem}
\newtheorem{lemma}{Lemma}
\newtheorem{proposition}{Proposition}

\newtheorem{remark}{Remark}

\begin{document}
\title{Cognitive Action Laws: \\ The Case of Visual Features}
%
%

\author{Alessandro~Betti
  \thanks{A. Betti is with University of Florence, Florence, Italy and DIISM, University of Siena, Siena, Italy.}, Marco~Gori,
  and~Stefano~Melacci
  \thanks{M. Gori and S. Melacci are with the DIISM, University of Siena, Siena, Italy.}
}


%
\maketitle              
\begin{abstract}
This paper proposes a theory for understanding perceptual learning processes within the general
framework of laws of nature. Neural networks are regarded as systems whose connections
are Lagrangian variables, namely functions depending on time. 
They are used to minimize the cognitive action, an appropriate functional index 
that measures the agent interactions with the environment. The cognitive action
contains a potential and a kinetic term  that nicely resemble the classic  
formulation of regularization in machine learning. A special choice
of the functional index, which leads to forth-order differential equations---Cognitive Action Laws (CAL)---exhibits a structure that mirrors classic
formulation of machine learning. In particular, unlike the action of  mechanics, the stationarity condition corresponds with 
the global minimum. Moreover, it is proven that typical asymptotic learning conditions
on the weights can coexist with the initialization provided that the system dynamics is
driven under a policy referred to as information overloading control.   
Finally,  the theory is experimented for the problem of feature extraction in computer
vision. 
\end{abstract}

%
%
%

\maketitle

\section{Introduction}
In machine learning one is typically involved in the longstanding dilemma on whether 
to use on-line or batch-mode learning. 
Clearly, the trend towards on-line  schemes is strongly motivated by the need
of learning on  huge training sets, though the forgetting behavior of stochastic
gradient descent algorithms have  promoted intermediate solutions based
on weight updating on mini-batches of data. 
When focussing on perceptual tasks, 
it is worth mentioning that the  iteration steps in on-line weight 
updating algorithms do not fully capture the notion of time. Whenever time does matter,
appropriate computational models have been proposed with the purpose of 
capturing temporal dependencies (see e.g. hidden Markov models and  
recurrent neural networks). In these cases, the relation between the iteration step
of the learning algorithm operating on sequences and time is quite involved 
and very much depend on the specific approach that has been adopted
(see e.g. the differences between recurrent networks with and without 
state relaxation towards a fixed point~\cite{Williams89b,Williams89c,Scarselli:2009:GNN}. 
A close relation between time and  neural activations is typically assumed
in neurobiological models, since in this case one is directly involved with
natural processes. Recently, there has been a growing interest in 
the formulation of computational models of learning that are fully embedded with time.  
When addressing the rising concerns on the biological plausibility of 
the backpropagation algorithm, in~\cite{scellier2018extending}, an extension of  
the framework of ``equilibrium propagation'' has been proposed that
does not compute the true gradient of the objective function~\cite{DBLP:journals/ficn/ScellierB17,scellier2018extending}. 
Using  statistical physics,  it has been shown that the inference processes from our sensory inputs and learning regularities  can be described by the same principles~\cite{Frinston2007}. The perceptual processes turn out to be an emergent property of systems that conform to a free-energy principle. Also in this case, the corresponding computational models are truly embedded in time, which leads to an extension of the gradient heuristics.

The approach followed in this paper is based on a formulation of learning that 
parallels the principle of least action in analytic mechanics, where the 
potential energy is related to the loss function, while a generalized
form of the kinetic energy is used to model the temporal evolution of the 
model parameters (weights)~\cite{Betti:2016:PLC}. 
While the proposed principle draws interesting paths to explore, there are at least a 
couple of remarkable issues that need to be addressed. 
First, unlike mechanics, it is clear that the
corresponding variational index, that is referred to as the cognitive action, needs to be
minimized. The authors of \cite{Betti:2016:PLC} did not stress minimization issues, thus fully 
paralleling the approach used in mechanics, where one is only looking for stationary points
of the action. While the proposed energy balancing approach somewhat helps understanding
the dynamics, there was no effort to discover the minimum of the action, which is
an important requirement for deeply understanding learning processes. 
Second, a more serious shortcoming in~\cite{Betti:2016:PLC} 
is that the concrete interpretation of the learning processes
driven by the Euler-Lagrange equations derived from the cognitive action requires the
fulfillment of boundary conditions, that are typically violated when following the inspiration 
from analytic mechanics, where one drives the trajectory from Cauchy's initial conditions. 
This is a serious problem, since we need a 
causal dynamic computational model to provide a truly on-line update of the parameters and, at the
same time, we need to satisfy the consistency with the boundary conditions.
The causality of the model is in fact required to gain the computational tractability, since any algorithmic 
search for the satisfaction of the boundary conditions seems to be hopeless.
The typical assumption behind learning policies is that of discovering constant weights at the
end of the learning process, which corresponds with imposing that all the temporal derivatives of the 
weights are null.

In this paper we address both limitations. First, we provide 
sufficient conditions for achieving the minimum of the cognitive action. 
As it will be shown, unlike mechanics, this requires to choose an action where the kinetic energy and
the potential comes with the same sign. This confers kinetic energy the role of classic 
regularization terms in machine learning.   
Second, we solve the problem of making the boundary conditions consistent with
Cauchy initialization, so as the Euler-Lagrange equations turn out to be a causal computational model. 
This is made possible by enforcing special dynamics over a manifold 
that corresponds with trajectories that are driven by an appropriate manipulation of
the input. In particular, we give conditions such that when the input is turned to zero
then all the temporal derivatives of the weights are quickly reset, thus respecting the 
boundary condition on the right border. Basically the model turns out to be strongly stable
which supports the need of quick dynamics.  
The intuition behind this solution comes from the principle of avoiding information overload, which
is somewhat similar to related ideas where the agent is expected to be exposed to the environment
according to a certain teaching plan (see e.g.~\cite{BengioICML2009},\cite{gori2016semantic}).

A fundamental result in this paper is that
the analysis on the minimality of the action joint with the need of enforcing stability 
leads to a choice of the kinetic terms that yields a forth-order associated
Euler-Lagrange differential equations of learning and inference, that
throughout the paper,  are referred to as the Cognitive Action Laws (CAL).  
The forth order turns out to be the minimum that guarantees the above conditions, 
which confers a special meaning to the special structure of  CAL presented in this paper. 

The proposed theory offers a framework to grasp 
an in-depth understanding of the dynamics of learning processes that 
are related to stochastic gradient to which they reduce under
an appropriate selection of the action parameters.
After having properly framed the CAL equations into the discrete setting of computation,
we carry out a preliminary experimental analysis of the theory for problems of visual feature extraction. 
The purpose of this analysis is not that of addressing classic computer vision benchmarks, but
is that of providing an experimental assessment of the novel concepts introduced in the theory 
in a specific example. 
In particular, we introduce an unsupervised learning process, that is based on the 
maximization of the mutual information from the video signal to a set of symbols. 
Basically, the mutual information turns out to be the potential of the action, while the
kinetic term drives the temporal evolution. The results show that the theory 
leads to construct a consistent unsupervised scheme whose features
resemble typical feature extraction in convolutional neural networks. This
opens the doors to a systematic adoption of the theory, where the agent is
simply exposed to its own visual environment. In a sense, this is a new perspective
in which one can think of learning of ``living agents'' whose behavior is driven 
by information-based laws. 


%
%

\section{Cognitive action}
\label{CA-lab}
Human cognitive processes do not present a neat distinction between training and test set. 
As time goes by, humans react surprisingly well to new stimuli, 
which suggests us to look for alternative foundations of learning
by embedding the agent into its own learning environment, so as 
we can think of learning as the outcome of laws of nature. 
This view of learning relies on the principle that the acquisition of cognitive skills 
obeys to information-based laws, which hold regardless of biology.
Based on~\cite{Betti:2016:PLC}, we establish a link with mechanics 
by paralleling the weights of a neural network
to the Lagrangian coordinates of a system of particles.
For reasons that will become clear in the rest of the paper, 
given an agent in its own environment, the  following functional, referred to as the cognitive action, 
will be used to drive inferential and learning processes: 
\begin{equation}\Gamma(q):=
  \int_0^T e^{\theta t}\Bigl(\xi R(q(t),\dot q(t),\ddot q(t))
  +  U\bigl(q(t),u(t)\bigr)\Bigr)\, dt.
\label{cognitive-action-1}\end{equation}
Here $\theta>0$, $q\colon \R^+\to \R^n$, $n\in \N$ are the weights of the learning system that are
the coordinates of a Lagrangian function composed of a regularization term\footnote{Throughout
the paper $|\cdot|$ is used to denote the norm.} 
$R(q,\dot q,\ddot q)=\xi k/2|q|^2+T(\dot q,\ddot q)$
and of a potential term $U\in C^1(\R^n\times\R^m)$, where $\xi \in \{-1,+1 \}$.
In particular, we choose  the following kinetic term
\[T(\dot q,\ddot q):=\frac{\alpha}{2} |\ddot q(t)|^2
  +\frac{\beta}{2}|\dot q(t)|^2+\frac{1}{2}|\gamma_1\dot q(t)
  +\gamma_2\ddot q(t)|^2 
\]
that somewhat characterizes the presence of an ongoing
learning process. 
The choice of this kinetic energy is related to
the one adopted in~\cite{Betti:2016:PLC}. As it will be shown in the remainder of the paper, 
the incorporation of second-order derivatives turns out to be useful  
when considering the need of minimizing the cognitive action while enforcing stability 
in the associated Euler-Lagrange equations. 
Notice that the kinetic energy is a sort of temporal regularization term that, once
minimized, leads to develop weights that settle to constant values, while the quadratic term
$\xi k/2  |q|^2$ is the classic regularization term that favor solutions with small weights.
The potential $U\in C^1(\R^n\times\R^m)$ depends on the function
$u\colon\R^+\to\R^m$, which describes the input information coming from the interactions with
the learning environment.  The underlying assumption is that $U(q,u) \geq 0$.
For example, in computer vision, $u$ is the video signal from which
the agent is expected to learn. The purpose of learning is that of achieving conditions under which,
for $t>\overline{t}$, we have $U(q(t),u(t)) \simeq 0$. Hence, according to machine learning
terminology, the potential can be regarded as a loss function. It is worth mentioning that
if $\xi=-1, \alpha=\gamma_{1}=\gamma_{2}=0$ then the cognitive action reduces to the 
classic action of analytic mechanics, where the regularization term $\xi k/2  |q|^2$ is aggregated with
the potential. Interestingly, in this case, one can look for learning processes that turn out to 
be a stationary point of the action, while they are not necessarily minima. 
Unlike~\cite{Betti:2016:PLC}, here, we are mostly concerned
with the extended notion of action, where we want to discover minima configurations.
Overall, the Lagrangian (\ref{cognitive-action-1}) contains the factor $e^{\theta t} \propto e^{\theta (t-T)}$ that provides 
a growing weight  as time goes by. The term $e^{\theta (t-T)}$ 
is equivalent under re-scaling of the cognitive action, and clearly prescribes that
the weight $1$ is gained at the end, while past information is gradually forgotten.
A clear motivation for this weighing will be given the in the following, but one can easily noticed
that it is a sort of discount factor that leads to focus more on recent information.  
If we pose $\mu=\alpha+\gamma_2^2$, $\nu=\beta+\gamma_1^2$, $\gamma=\gamma_2\cdot\gamma_2$, and $\xi=1$, then
Eq.~(\ref{cognitive-action-1}) can be rewritten as
\begin{equation}\Gamma(q)=
  \int_0^T e^{\theta t}\Bigl(\frac{\mu}{2} |\ddot q|^2
  +\frac{\nu}{2}|\dot q|^2+\gamma\dot q\cdot\ddot q
  +\frac{k}{2}|q|^2
  +U\bigl(q,u\bigr)\Bigr)\, dt.\label{cognitive-action-2}\end{equation}
The interpretation of learning by means of functional~(\ref{cognitive-action-2})
is especially interesting since, unlike the case of the classic action in mechanics, 
it admits a minimum under appropriate conditions. Of course, this property makes it more
attractive for machine learning. 
\begin{theorem}
	If the following coercivity conditions
	\begin{equation}
          \mu>\gamma_{2}^{2},\quad \nu>\gamma_{1}^{2},\quad k>0
          \label{coerc-cond}
	\end{equation}  
	hold true then functional $\Gamma$,  defined by Eq.~\ref{cognitive-action-2},
	admits a minimum on the set
        \[\mathscr{K}=\{\,q\in H^2((0,T), \R^n)\mid q(0)=q^0, \dot q(0)=q^1\,\}.\]
\label{GlobalMinimaRes}
\end{theorem}
\begin{proof}
See Appendix A.
\end{proof}
%
We report a few qualitative comments in order to understand this result.
First, we notice that, unlike mechanics, the choice $\xi=1$ helps the lower
boundedness of $\Gamma$. This is immediately clear when $\gamma=0$,
but the term $\dot{q} \ddot{q}$ will be proven to 
play a fundamental role for the approximation of stochastic gradient dynamics. 
Its sign  contributes to develop solutions that generate consistent learning trajectories
while minimizing $\Gamma(q)$.
Suppose the weights are growing, that is $\dot{q}>0$. Then a trajectory
following $\ddot{q}<0$ yields a concave function that clearly contributes to minimize 
$\Gamma$. Likewise, the same holds for $\dot{q}<0$ and $\ddot{q}>0$ 
acts coherently while enforcing convexity of $q$. At the light of these comments,
the coercivity conditions~(\ref{coerc-cond}) clearly contribute to impose 
a lower bound on $\Gamma$. The reason is that the correspondent choices of $\mu$ and $\nu$ 
leads attribute a relevant weight to the second- and first-order kinetic terms that, 
unlike $\gamma \dot{q} \ddot{q}$, are positive. 

In order to determine the minimum, we must impose the conditions for determining stationary 
points, which require the fulfillment of the  Euler-Lagrange equations. 
We will perform the variation of $\Gamma$
in the  general assumption that $\mu$, $\nu$,
$\gamma$, $k$ and $U$ have an explicit dependence on time. Although this
does not change the structure of the resulting differential equation, it will
turn out to be useful in the remainder of the paper. 
In order to simplify the calculations of the variation 
we use the following equivalent expression of $\Gamma$: 
\begin{equation}
  \begin{split} \Gamma(q)=\mskip -3mu
  \int_0^T \Bigl(&\frac{\hat \mu(t)}{2} |\ddot q(t)|^2
  +\frac{\hat \nu(t)}{2}|\dot q(t)|^2+\hat \gamma(t)\dot q(t)\cdot\ddot q(t)\\
  &+\frac{\hat k(t)}{2}|q(t)|^2
  +\hat U\bigl(t,q(t),u(t)\bigr)\Bigr)\, dt;
  \end{split}\label{cognitive-action-3}\end{equation}
 where $\hat\mu(t)=e^{\theta t}\mu$,
$\hat\nu(t)=e^{\theta t}\nu$, $\hat\gamma(t)=e^{\theta t}\gamma$,
$\hat k(t)=e^{\theta t}k$, and $\hat U(t,q(t),u(t))=e^{\theta t}U(q(t),u(t))$.
Now, let us consider the variation $v$ and define 
$\psi(s)=\Gamma(q+sv)$, where $s \in \R$.
In the analysis below, we will repeatedly use the fact
that $v(0)=\dot v(0)=0$. This corresponds with the assignment of the initial values  
$q(0)$ and $\dot{q}(0)$. Since we want to provide a causal computational framework
for $q(t)$, this is in fact the first step towards this direction. 
The stationarity condition for the
functional $\Gamma$ is $\psi'(0)=0$, \footnote{Here and in the rest of the paper, we sometimes simplify the notation by removing the explicit dependence on time.}
\[
  \begin{split}
\psi'(0)=&\big[(\hat\mu\ddot q+\hat \gamma\dot q)\dot v+
\big(\hat\nu\dot q+\hat\gamma\ddot
q -(\hat\mu\ddot q+\hat \gamma\dot q)\dot{ }\big)v\big]_{t=T}
\\
&+\int_0^T \big\{(\hat\mu\ddot q+\hat \gamma\dot q)\,\ddot{ }\,
- \big(\hat\nu\dot q+\hat\gamma\ddot
q\big)\dot{ }\, +\nabla_q \hat U\big\}\cdot v=0.
\end{split}
\]
According to the fundamental lemma of variational calculus, 
if $q$ is a minimum, the above expression should hold for all the
allowed variations $v$, so as we can proceed as follows:
\begin{enumerate}
\item Consider only the variations such that $v(T)=\dot v(T)=0$. In this case
  $\psi'(0)=0$ yields the following differential equations
  \begin{equation}
    \hat\mu q^{(4)}+2\dot{\hat\mu} q^{(3)}+(\ddot{\hat \mu}+\dot{\hat
\gamma}-\hat \nu)\ddot q +(\ddot {\hat \gamma}-\dot{\hat\nu})\dot
q+\hat k q+\nabla_q \hat U=0.
\label{CAL-eq}
\end{equation}
\item Because of Eq.~(\ref{CAL-eq}), $\psi'(0)=0$ reduces to
  $\big[(\hat\mu\ddot q+\hat \gamma\dot q)\dot v+
\big(\hat\nu\dot q+\hat\gamma\ddot
q -(\hat\mu\ddot q+\hat \gamma\dot q)\dot{ }\big)v\big]_{t=T}=0$. Moreover,
since $v(T)$ and $\dot v(T)$ can be chosen independent one of each other, then
the vanishing of the first variation also implies that
\begin{equation}
\begin{split}
  &\hat\mu\ddot q(T)+\hat\gamma\dot q(T)=0;\\
  &\hat\mu q^{(3)}(T)+\dot{\hat\mu}\ddot q(T)+(
  \dot{\hat\gamma}-\hat\nu) \dot q(T)=0.
 \end{split}\label{neumann}
\end{equation}
\end{enumerate}
The set of triples $(\dot{q}(T),q^{(2)}(T),q^{(3)}(T))$ that satisfy this condition 
is denoted by $\mathscr{N}_{T}$.
Now, if we restore the explicit dependence of the coefficients on the term
$e^{\theta t}$ then Eq.~(\ref{CAL-eq}) reads
\begin{equation}
  \begin{split}
  \mu q^{(4)}+2\theta\mu q^{(3)}+(\theta^2\mu+\theta\gamma-\nu)\ddot q
  &+(\theta^2\gamma-\theta\nu)\dot q\\ &+ kq+\nabla_q U(q,u)=0.
  \end{split}
\label{CAL-eq-2}
\end{equation}
These equations are referred to as the Cognitive Action Laws (CAL) of learning.
Notice that the eventual non-linearity in this differential
equation resides entirely in the gradient term; the remaining part is a linear ODE. 
When joining the result stated by Theorem~\ref{GlobalMinimaRes} and the above analysis, 
we can state the following theorem:
\begin{theorem}
If the weight function $q^{\star} \in \mathscr{K}$
that satisfies the CAL equations~(\ref{CAL-eq-2})  is consistent with the right-boundary
conditions~(\ref{neumann}) then it is the argument of the minimum of $\Gamma(q)$.
\label{EL-CAL-eq}
\end{theorem}
Notice that the two initial conditions on $q(0)=q^0, \dot q(0)=q^1$, along with 
the right-boundary conditions~(\ref{neumann}), guarantee the existence of the
minimum. While this is an interesting result, unfortunately, 
Theorem~\ref{EL-CAL-eq} does not offer a direct computational scheme
for determining the minimum. There is in fact the typical causality issue that
arises whenever one wants to optimize over time. 
Basically, the additional Neumann-like
conditions~(\ref{neumann}) may be in conflict with the Cauchy conditions to be
used for a causal solution. As pointed out in the introduction, this was in fact
one of important problem left unsolved in~\cite{Betti:2016:PLC}.
In the next section, however, we will show that a careful modulation of the input signal
$u(t)$, that does not change the nature of the learning and inference task, 
allows us to establish a causal computational scheme that verifies, 
with an arbitrary degree of precision, the conditions~(\ref{neumann}).


\section{Boundary conditions and developmental issues}
%
%
\label{sec:bound}
The  theoretical results of the previous section suggest to formulate learning and inference 
as the problem of determining 
\begin{equation}
	q^{\star}(t) = \argmin_{q \in \mathscr{Q}} \Gamma(q),
\label{LearnDef}
\end{equation}
where the functional space  $\mathscr{Q}$ is constructed from $\mathscr{K}$ 
with augmented left and right boundary conditions:
\[\mathscr{Q} = \{\,q\in \mathscr{K}\mid 
(\dot{q}(T),q^{(2)}(T),q^{(3)}(T)) \in \mathscr{N}_{T}\}.
\] 
Clearly, we cannot search in this space by a causal
computational scheme, which would require to set also the values of $q^{(2)}(0)$
and $q^{3}(0)$. These two additional conditions lead in fact to an overdetermined
set of equations for finding the coefficients of the CAL equations. 
Hence, the chosen Cauchy conditions likely violates the Neumann-like
conditions~(\ref{neumann}). It is worth mentioning that this degree of violation
likely depends on $T$, and that if $T \rightarrow 0$ the right-boundary conditions
collapses to the Cauchy conditions (in the case of null initial conditions). 
An early discussion on how to make causality consistent with 
Neumann-like conditions~(\ref{neumann}) was given in~\cite{DBLP:journals/corr/abs-Betti2018}. As discussed in~\cite{MarcoGori-MK-2018} (Ch. 6), the  basic idea can 
be naturally framed in the context of developmental learning, according to which 
one does not overload the agent with all the available information, but filter it properly
so as to gain a causal optimization of $\Gamma$. In other words, the environmental
information is presented gradually to the agent so as to favor the acquisition of
small chunks, for which the approximate satisfaction of the boundary conditions
is facilitated. The gradual exposition of the agent can also benefit from an 
appropriate filtering of the input with the purpose of reducing the associated 
information. 
In experimental results reported in Section~\ref{sec:visexp} we will provide evidence
that  a suitable modulation of the input signal, that does not change the nature of
the problem, allows us to  solve
Eq.~(\ref{CAL-eq}) with Cauchy conditions in such a way that
the boundary conditions~(\ref{neumann}) are satisfied.
%
%
%

Now we show that there is another way of controlling the information overloading.
Intuitively, the consistency with the boundary conditions can be gained
by two decoupled dynamics in Eq.~(\ref{CAL-eq}), one of which
performs a ``reset'' of the derivatives. 
Hence, the divergent dynamics that gives rise from the Cauchy's initialization 
and from the natural rhythm of incoming information can be controlled by
discharging the learning state accumulated in the weights.
While such a decoupling dynamics is admissible in the general model
of CAL equations, we will show that we can always choose the parameters
in such a way to implement the ``reset''. Like in the previous discussion 
on developmental issue, where the information is supposed to be gradually presented, 
the rationale behind this result is that learning processes are typically consistent with 
the temporary detachment of the input $(u=0)$ for arbitrarily small intervals.
Consider a sequence
of times $0<t_0<t_1<t_2<\cdots<t_{2N}<T$ that defines the two sets
$A=\bigcup_{i=0}^{N}A_i$ with $A_i=(t_{2i-1},t_{2i})$, $t_{-1}=0$
and $B=\bigcup_{i=0}^{N} B_i$ with $B_i=(t_{2i},t_{2i+1})$, $t_{2N+1}=T$
and we let
\footnote{We use Iverson's notation:
 Given a statement $A$ we set $\(A)$ to $1$ if $A$ is true and to
$0$ if $A$ is false.}
\begin{equation}\begin{split}
  &\hat\mu(t)=e^{\hat \theta(t) t}
  \bigl(\mu\, \(t\in A)+\bar\mu\, \(t\in B)\bigr);\\
  &\hat\nu(t)=e^{\hat \theta(t) t}
  \bigl(\nu\, \(t\in A)+\bar\nu\, \(t\in B)\bigr);\\
  &\hat\gamma(t)=e^{\hat \theta(t) t}
  \bigl(\gamma\, \(t\in A)+\bar\gamma\, \(t\in B)\bigr);\\
  &\hat k(t)=e^{\hat \theta(t) t}
  \bigl(k\, \(t\in A)+\bar k\, \(t\in B)\bigr),\cr
\end{split}\label{coeff-in-time}\end{equation}
with $\hat\theta(t)=\theta\, \(t\in A)+\bar\theta\, \(t\in B)$. Not only
Eq.~(\ref{CAL-eq}) still holds, but with this special temporal dependence,
the equations turn out to be decoupled,
for all times apart from $t_0,t_1,\dots,t_{2N}$, as follows:
\begin{equation}
  \begin{split}
   \mu q^{(4)}+2\theta\mu q^{(3)}+(\theta^2\mu&+\theta\gamma-\nu)\ddot q
   +(\theta^2\gamma-\theta\nu)\dot q\\
   &+ kq+\nabla_q U=0, \,  t\in A;
   \end{split}\label{eq_tA}
\end{equation}
and
\begin{equation}
  \begin{split}
   \bar\mu q^{(4)}+2\bar\theta\bar\mu q^{(3)}+(\bar\theta^2\bar\mu
&+\bar\theta\bar\gamma-\bar\nu)\ddot q
+(\bar\theta^2\bar\gamma-\bar\theta\bar\nu)\dot q\\
&+ \bar kq+\nabla_q U=0,
\, t\in B.
\end{split} \label{eq_tB}
\end{equation}
We make the fundamental assumption of controlling the input
by $u(t)\to u(t)\,\(t\in A)$, that is to reset the input when we are outside set $A$.
This choice comes when one bears in mind the previous discussion on the
need to fulfill the boundary conditions. As a consequence
the equation that describes the temporal evolution for $t\in B$ reduces to the linear equation
\begin{equation}
  \bar\mu q^{(4)}+2\bar\theta\bar\mu q^{(3)}+(\bar\theta^2\bar\mu
+\bar\theta\bar\gamma-\bar\nu)\ddot q
+(\bar\theta^2\bar\gamma-\bar\theta\bar\nu)\dot q+ \bar kq = 0,
\, t\in B.
\label{dyn-in-B}
\end{equation}
Now we state two important theorems that
show how the input $u(t)\to u(t)\,\(t\in A)$ leads to matching the desired
boundary conditions thus gaining the 
consistency with Cauchy initialization.
\begin{theorem}
  We can choose the system parameters of Eq.~(\ref{dyn-in-B}) 
  in such a way that  $|q^{k}(t_{2i+1})|=0$ 
  up to an arbitrary precision for  $i=0,1,\dots, N$ 
  regardless of the initial Cauchy  conditions,
  which is in fact a special way of satisfying boundary conditions~(\ref{neumann}).
   \label{vanishing-derivatives}
\end{theorem}
\begin{proof}
	See Appendix A.
\end{proof}

While this theorem guarantees the consistency between Cauchy's
initialization and the boundary conditions, one might wonder whether
the reset of the derivatives of the weights in any segment $B_{i}$ 
can also be paired with the latching of the weights developed in the previous
segment $A_{i}$.
As stated in the following
theorem the choice of the roots of the characteristic polynomial
in Eq.~(\ref{dyn-in-B}) 
guarantees that the values of $q(t)$ at the beginning and at the end
of any time interval $B_i$, $i=0,\dots, 2N$ is the same
with arbitrary degree of precision. We will show that if we choose the roots 
$(0,\lambda_{2},\lambda_{3},\lambda_{4})$, then the 
value $\rho=\max_{k}|\lambda_k|$ to achieve a precision $\epsilon$  depends
on the Vandermonde matrix $V_{3}=V(\lambda_{2},\lambda_{3},\lambda_{4})$,
on the value of the derivatives $\dot q$, $\ddot q$ and $q^{(3)}$
at $t_i$ on $\epsilon$ and on a suitable constant $C$ that bounds the entries of the
inverse of the Vandermonde matrix $V(\lambda_2/\rho,\lambda_3/\rho,\lambda_4/\rho)$).
%
\begin{theorem} 
	Let $\Lambda=(V(\lambda_2/\rho,\lambda_3/\rho,\lambda_4/\rho))^{-1}$ be.
  	For every even $i=0,\dots, 2N$ consider the defined sets
  	$A_i=(t_{i-1}, t_i)$, $B_i=(t_i, t_{i+1})$.
  	It is always possible to choose the coefficients in Eq.~(\ref{dyn-in-B})
  	such that $\forall \epsilon>0$, if we choose 
	\[
		\rho > [(9C/\epsilon)  \cdot\max_k |q^{(k)}(t_i)|]^{1/2} >1
	\]
         we have $|q(t_{i+1})-q(t_i)|<\epsilon$, 
         where  $|\Lambda_{kj}|\le C$ for all $k$ and $j=1,2,3$.
\label{MemoryDNS}
\end{theorem}
\begin{proof}
	See Appendix A.
\end{proof}
This theorem enables the replacement of  the solution of Eq~(\ref{dyn-in-B})
with the enforcement of a reset as it is described in Section~\ref{sec:visexp}.
Basically, the information overloading associated with the temporal presentation of
the source can be properly controlled by resetting all derivatives of the weights, while
keeping their value. The corresponding solution keeps all the discussed properties 
and, particularly, makes Cauchy initialization consistent with the boundary conditions, an
issue which was left open in~\cite{Betti:2016:PLC}. The conclusion that can be drawn from  
Theorems~\ref{vanishing-derivatives} and~\ref{MemoryDNS} is quite surprising, since the
reset of the derivatives turns out to be fully consistent with the causality of the problem.
%


\section{CAL dynamics}
\label{sec:dyn}
In this section we discuss the dynamics behind the cognitive action
laws stated by Eq.~(\ref{CAL-eq-2}). This is important for the appropriate set up
of the parameters in the application to any cognitive task, like the one of vision 
described in the following experimental section.
In particular,  we will focus on case in which $u\equiv0$, where the 
free dynamics is driven by the kinetic term only. 
We also address the relationships of CAL dynamics with classic stochastic gradient and
prove that it can be reproduced under appropriate choices of the parameters.

\subsection{Free dynamics}
On null input, since we  assume that $U(q,0)=0$, Eq.~(\ref{CAL-eq-2})
becomes $q^{(4)}+b q^{(3)}+c \ddot q+d \dot q+ eq=0$,
where we assume $\mu\ne 0$ and use the notation $b=2\theta$,
$c=(\theta^2\mu+\theta\gamma-\nu)/\mu$, $d=(\theta^2\gamma-\theta\nu)/\mu$,
and $e=k/\mu$. The solution is fully characterized by the nature of the roots of the characteristic
polynomial
$\chi(x)=x^4+bx^3+cx^2+dx+e.\label{char-pol}$
In particular the behavior of the solution
is mainly affected by the negativeness of the real part of the roots and
by their imaginary part.
The first condition ensure the asymptotic stability of the solution, while the
violation of the second one prevents oscillatory behavior.

\begin{lemma}
  The characteristic polynomial $\chi(x)$ with real coefficients
  is asymptotically
  stable if and only if
  \begin{equation}
    b>0,\quad c>0,\quad 0<d< bc, \quad 0<e<\frac{bcd-d^2}{b^2}.
    \label{stab-cond}
    \end{equation}
 	\label{StabLemma}
  \end{lemma}
  \begin{proof}
    The proof is gained by the straightforward application the Routh-Hurwitz criterion (see for
    example~\cite{uspensky1948theory}). 
    \end{proof}
    If we replace $x=z-b/4$ with $\chi(x)$ then
    we obtain the reduced quartic equation
    $\zeta(z):=\chi(z-b/4)=z^4+q z^2+r z+s=0,$
where
$q=c-3 b^2/8, \quad r=b^3/8-bc/2+d,\quad s=
b^2c/16-3/256 b^4-bd/4+e$.
    
\begin{lemma}
  The characteristic polynomial $\zeta(z)$
  with real coefficients $q,r,s$ and with discriminant $\Delta$ has
  only real roots if:
  \begin{enumerate}
  \item $q<0$, $4s-q^2<0$ and $\Delta>0$ (4 distinct real roots);
  \item $-q^2/12<s<q^2/4$ and $\Delta=0$ (roots real, only two equal);
  \item $q<0$, $s=q^2/4$ and $\Delta=0$ (two pair of equal real roots);
  \item $q<0$, $s=-q^2/12$ and $\Delta=0$ (all roots real, three equal);
  \item $q=0$, $s=0$ and $\Delta=0$.
  \end{enumerate}
  \label{reality-lemma}
\end{lemma}
\begin{proof}
  See~\cite{Rees1922}.
  \end{proof}

  \begin{proposition}
    If we choose $\theta, \mu, \nu, \gamma_1,\gamma_2, k$ such that
    $\theta>0$ and:
    \begin{align}
    \begin{split}
    	&\mu<\gamma_2^2,\quad \nu<\gamma_1^2,\quad \nu<\theta
      	\gamma_1\gamma_2,\quad
        0<k<\frac{(\nu-\theta\gamma_1\gamma_2)^2}{4\mu} \\
    	&\gamma_1<0,\quad\gamma_2<\frac{\gamma_1}{\theta}\quad \hbox{or}\quad
    \gamma_1>0,\quad\gamma_2>\frac{\gamma_1}{\theta}.
    \end{split}
    \end{align} 
    then the following conditions are jointly verified:
    \begin{enumerate}
    \item $\Gamma$ admits a minimum in $\mathscr{K}$;
    \item the homogeneous equation associated with Eq.~(\ref{CAL-eq-2}) has the following two 
    	properties:
    	\begin{enumerate}
     	\item [$i.$] it is asymptotically stable;
    	\item [$ii.$]it yields aperiodic dynamics  (the roots of the characteristic polynomial are real).
	\end{enumerate}
      \end{enumerate}
      \label{prop:coef}
\end{proposition}
\begin{proof}
  The proof follows of $1$ on the admission of a minimum 
  comes from Theorem~\ref{GlobalMinimaRes} when considering
  that, under the given assumptions, conditions~(\ref{coerc-cond}) hold true.
  As for the statements $2$, the proof of $i$ comes from Lemma~\ref{StabLemma} (stability) and
  from Lemma~(\ref{reality-lemma}) (aperiodicity).
  \end{proof}
\subsection{Reproducing gradient flow}
Let us consider Eq.~(\ref{CAL-eq-2}) with $\mu=\nu\equiv 0$ and
$\gamma=1/\theta^2$.  Then this equation reduces to
\begin{equation}
  \theta^{-1}\ddot q(t)+\dot q(t)=-kq(t)-\nabla_q U(q(t), u(t)).
\label{reduction-gradient}
\end{equation}
We can promptly see that 
as  $\theta\to\infty$ the CAL equation~(\ref{CAL-eq-2})
restores the classic gradient flow with potential $k/2 |q|^2+U(q,u)$.
As anticipated in Section~\ref{CA-lab}, gradient flow arises 
from the term $\dot{q} \ddot{q}$, whose intuitive contribution
to the system dynamics was already given.
The choice of the parameters that reduces CAL dynamics
to a gradient flow transforms the boundary conditions~(\ref{neumann})
into $\dot{q}(T)=0$. This is in fact the ordinary condition that one expect to be matched
at the end of gradient-driven learning processes, namely that the weights converge to a constant
value. Clearly, for such a convergence we tacitly assume that the learning task presents some
form of regularity to be induced. A recent result in this direction is given in~\cite{Bellettini-Betti-Gori-NIPS-2018}. 

Notice that gradient flow is also
recovered from the action of analytic mechanic
with strong dissipation. This correspond with choosing $\xi=-1$,
$\mu=\gamma=0$ and $\nu=1/\theta$. 
In this case the Euler Lagrange Equation  reduces to
\begin{equation}
	-\theta^{-1}\ddot q(t)-\dot q(t)-kq(t)-\nabla_q U(q(t),u(t))=0.
\end{equation}
Like Eq.~(\ref{reduction-gradient}), as $\theta\to\infty$, also the above equation indeed returns a gradient flow. Thus in both cases the
Euler approximation is $q(k+1)=q(k)-k q(k)-\nabla_qU(q(k),u(k))$.

The importance of the incorporation of gradient flow in CAL equations
is that some of the results can be inherited also for classic gradient 
descent algorithms that are massively used in most applications
of machine learning. On-line stochastic gradient and 
gradient descent on mini-batches are typically given a foundation
by they association with 
batch mode gradient to invoke a sort of overall minimization property
that emerges from data redundancy. The theory herein presented
offers a clear foundations of those intuitive connections
in a natural framework driven by the temporal representation
of the input in the context of variational calculus.

\section{The case of visual features}
\label{sec:visexp}

In this section we carry out an experimental analysis aimed at understanding the 
dynamics of CAL equations with the final purpose of exploring their behavior in 
learning tasks. We are mostly interested in understanding the role of the 
different parameters in the action functional and to validate the theoretical 
results stated in the previous section. In addition to the experimentation of
the causal processing scheme we aimed at verifying the important role of
filtering the input, as well as that of properly resetting the system dynamics. 
More specifically, we are not interested in carrying out experiments on classic
benchmarks, that are typically based on large image collections, but on 
checking the agent behavior on real-world visual environments. 
With this purpose in mind, 
we consider the problem of unsupervised learning of visual features from videos. 
It is a classic perceptual task where the role of time plays a crucial role.

Let $X$ be the set of pixel coordinates and denote by $u$ 
the input video, where $u(t)$ is the frame at time $t$. 
We extract $n$ convolutional features from each pixel, where the coefficients of the convolutional filters are stored into $q$. 
If the size of a filter is $f\times f$ (for each of the $m$ input channels), then the number of components of
$q(t)$ is $n\cdot f^2 \cdot m$.
The activations of the features on the pixel $x\in X$ at a certain time $t$ are given by $\Phi(x,t)=\sigma\bigl(q(t)\ast u(t)\bigr)$,
where $\sigma\colon\R^n\to\R^n$ is the softmax function 
and $q(t)\ast u(t)$ is the convolution of the video with the $n$ filters computed 
in the pixel of coordinates $x$.

A possible criterion to learn the filters $q(t)$ is to require that
the Mutual Information (MI) between the input video and the extracted features is maximized \cite{melacci2012unsupervised,gori2016semantic}.
Instead of using the Shannon entropy we use the quadratic entropy
$-\sum_i p_i^2$,  and the following associated potential can be chosen which play the same role as 
maximizing the MI\footnote{In this case it is always possible to
achieve $U(q,0)=0$ just by adding a constant.}: 
\begin{equation}
U(q,u):=\sum_{i=0}^{n-1}\biggl[{\lambda_E\over2}\biggl(\sum_{x\in X}
g_x \sigma\bigl(q\ast u\bigr)\biggr)^2
-{\lambda_C\over2}
\sum_{x\in X} g_x \Bigl(\sigma\bigl(q\ast u\bigr)
\Bigr)^2\biggr],
\label{MI}
\end{equation}
where $g_x$ is a probability measure on the pixel coordinates, that is supposed to be uniform. 
For an in-depth discussion on this equation see~\cite{DBLP:journals/corr/abs-Betti2018}.
Here, we we follow the spirit of MaxEnt and relax the definition of MI by introducing 
the  parameters $\lambda_E, \lambda_C>0$ that weigh the contribution of the two entropies. 




We implemented a solver for the CAL of Eq. (\ref{CAL-eq-2}) that is based on the Euler method with step size $h$. After having reduced the CAL equations to the first order, the variables that were updated 
at each $t$ are $q$, $\dot q$, $\ddot q$, and $q^{(3)}$. The code and data we used to run the following experiments can be downloaded at \url{https://github.com/alessandro-betti/see}, together with the full list of model parameters.
We randomly selected two real world video sequences from the Hollywood Dataset HOHA2 \cite{marszalek09}, that we will refer to as ``skater'' and ``car'', and a clip from the movie ``The Matrix'' (\textcopyright  Warner Bros. Pictures). The frame rate of all the videos is $\approx$ 25 fps, each frame was rescaled to $240\times110$ and, unless differently specified, converted to grayscale. Videos have different lengths, ranging from $\approx 10$ to $\approx 40$ seconds, and they were repeated in loop until $45,000$ frames were generated, thus covering a significantly longer time span. 
We randomly initialized $q(0)$  while the derivatives at time $t=0$ were set to $0$.
Following the developmental plan indicated in Section \ref{sec:bound}, the video was 
gradually presented to the system, starting from a completely null signal (all pixel intensities are zero), and slowly increasing the level of detail and the pixel intensities, in function of $\phi(t) \in [0,1]$,
$u(t) = \phi(t) \left( \texttt{gauss}\left((1-\phi(t)\delta)\right) \ast u^{o}(t) \right)$, 
where $u^{o}(t)$ is the source video signal, $\texttt{gauss}(\sigma^2)$ is a Gaussian filter of variance $\sigma^2$, and $\delta \in [0,1]$ is a customizable scaling factor. 
We start with $\phi(0)=0$, and then $\phi$ is progressively increased as time passes, $\phi(t+1) = \phi(t) + \eta(1-\phi(t))$ (we set $\eta = 0.0005$). 
We refer to the quantity $1-\phi$ as the ``blurring factor''.

According to the indications of Section~\ref{sec:bound}, we also 
carried out the ``reset plan'' according to which the video signal undergoes a reset
whenever the derivatives become too large. Formally, if $\|\dot q(t')\|^2\geq\epsilon_{1}$, or $\|\ddot q(t')\|^2\geq\epsilon_{2}$, or $\|q^{(3)}(t')\|^2\geq\epsilon_{3}$ then we forced $\phi(t')$ to $0$, 
switching from the case of Eq. (\ref{eq_tA}) to the one of Eq. (\ref{eq_tB}) ($\epsilon_j=300 \cdot n$, for all $j$), and then we set to $0$ all the derivatives. 

We evaluated the CAL dynamics  by experimenting four instances of the set of parameters $\{\mu, \nu, \gamma, k\}$. Each instance is characterized by the roots of the characteristic polynomial that lead to \textit{stable} or \textit{not-stable} configurations, and with only \textit{real} or also \textit{imaginary} parts, keeping the roots close to zero, and fulfilling the conditions of Proposition \ref{prop:coef} when stability and reality are needed. These configurations are all based on values of $k \in [10^{-19}, 10^{-3}]$, while $\theta=10^{-4}$.

\def\|#1|{\includegraphics[width=.5\hsize]{#1}}
\def\<#1>{\includegraphics[width=.3\hsize]{#1}}
\begin{figure*}
\tabskip=1em plus2em minus.5em
  \halign to\hsize{\hfil#\hfil&\hfil#\hfil&\hfil#\hfil\cr
  $\quad\hbox{FPS}=25,\quad s=1/25$&$\quad\hbox{FPS}=10\,\quad s=1/10$&$\quad\hbox{FPS}=25,
  \quad s=1/100$\cr
\noalign{\medskip}
  \<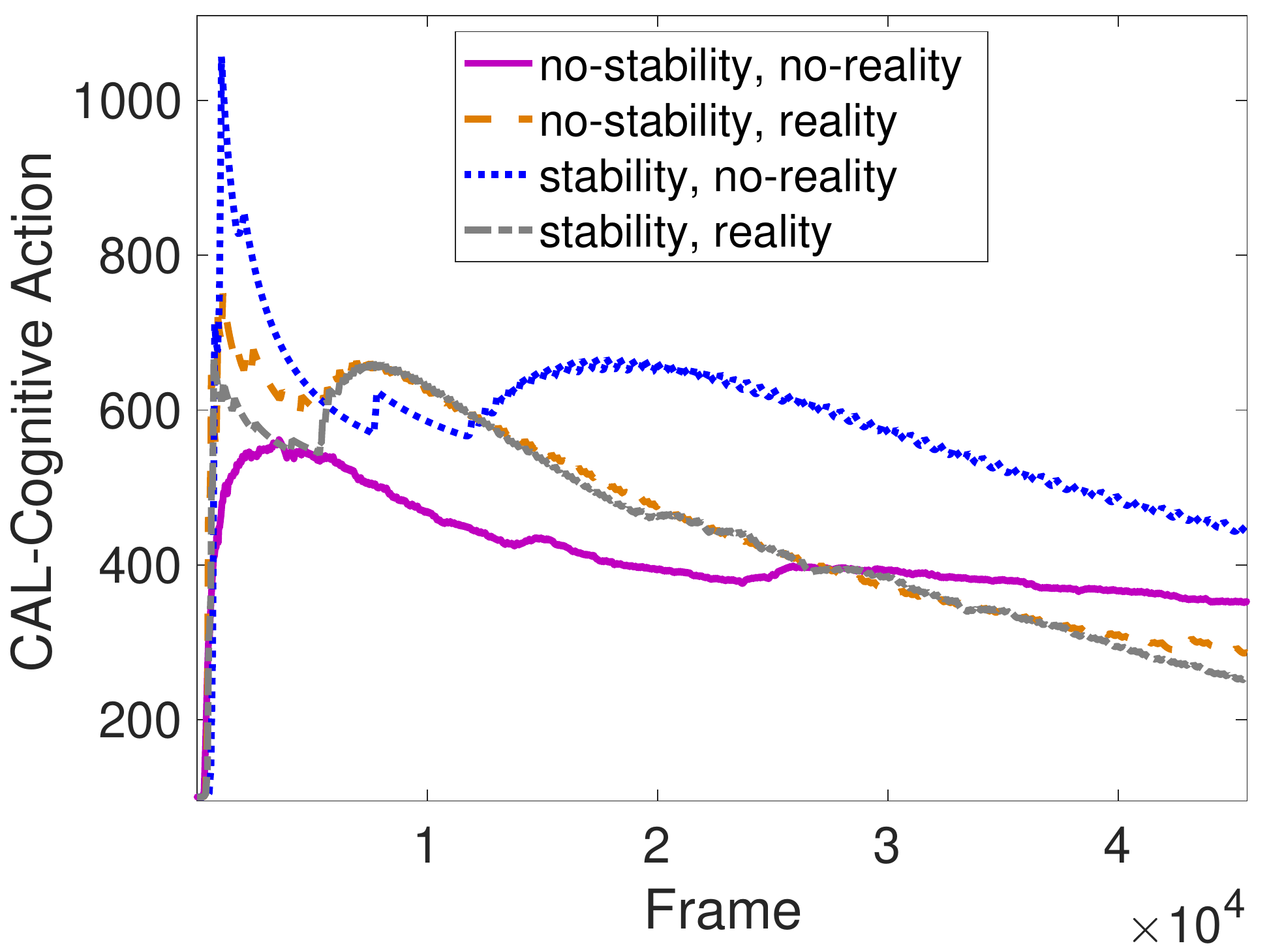>&\<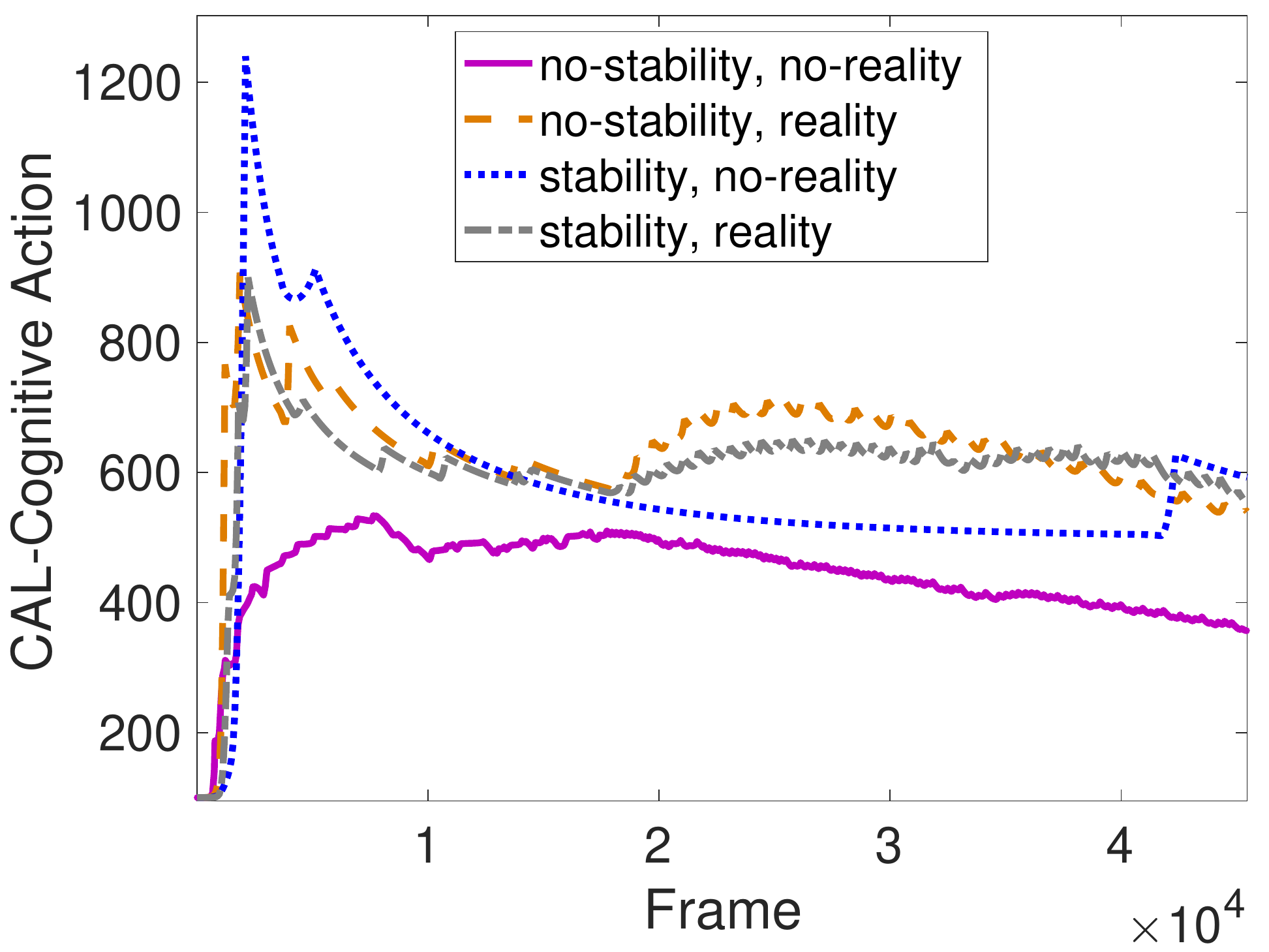>&
  \<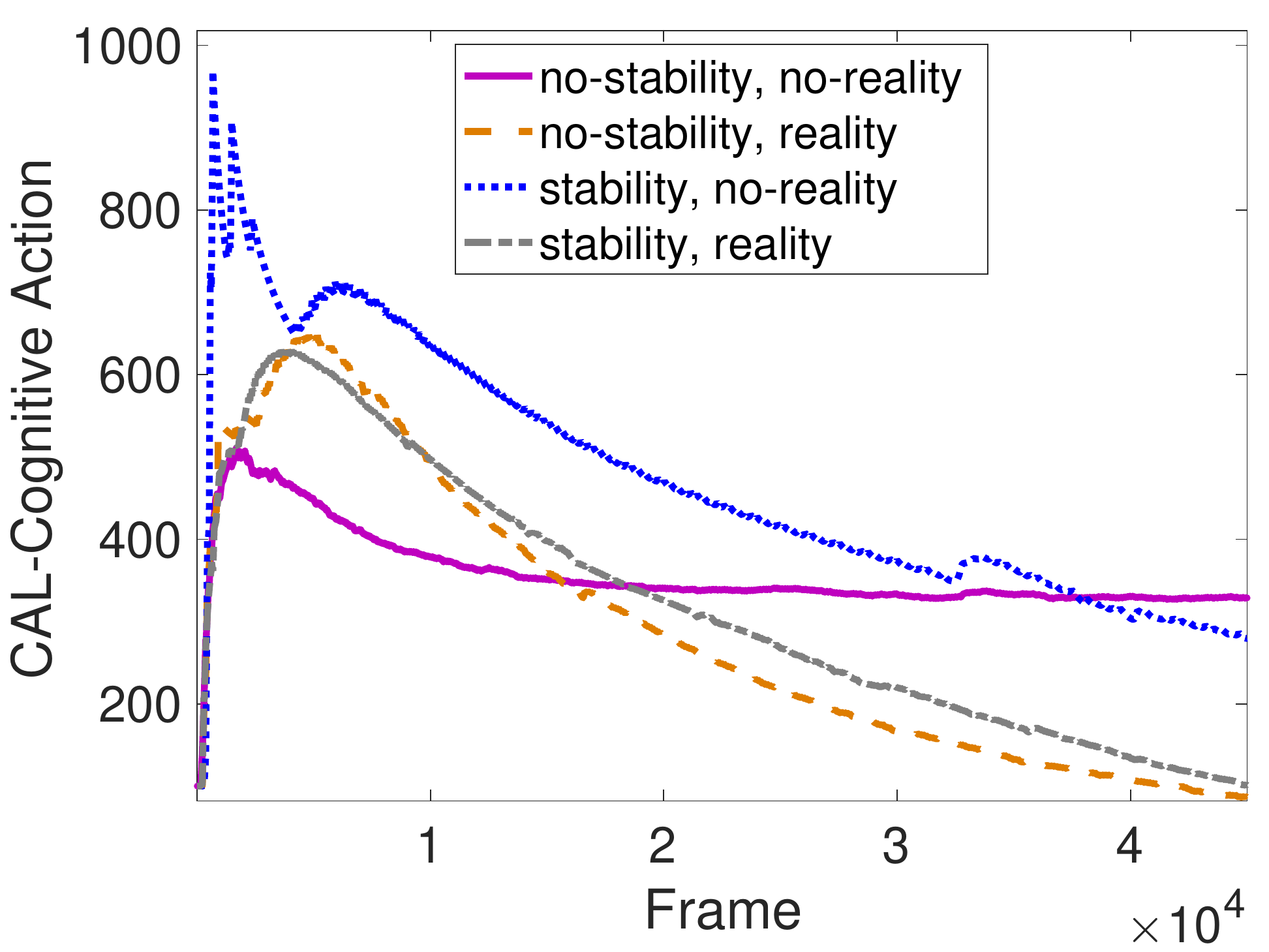>\cr
  \<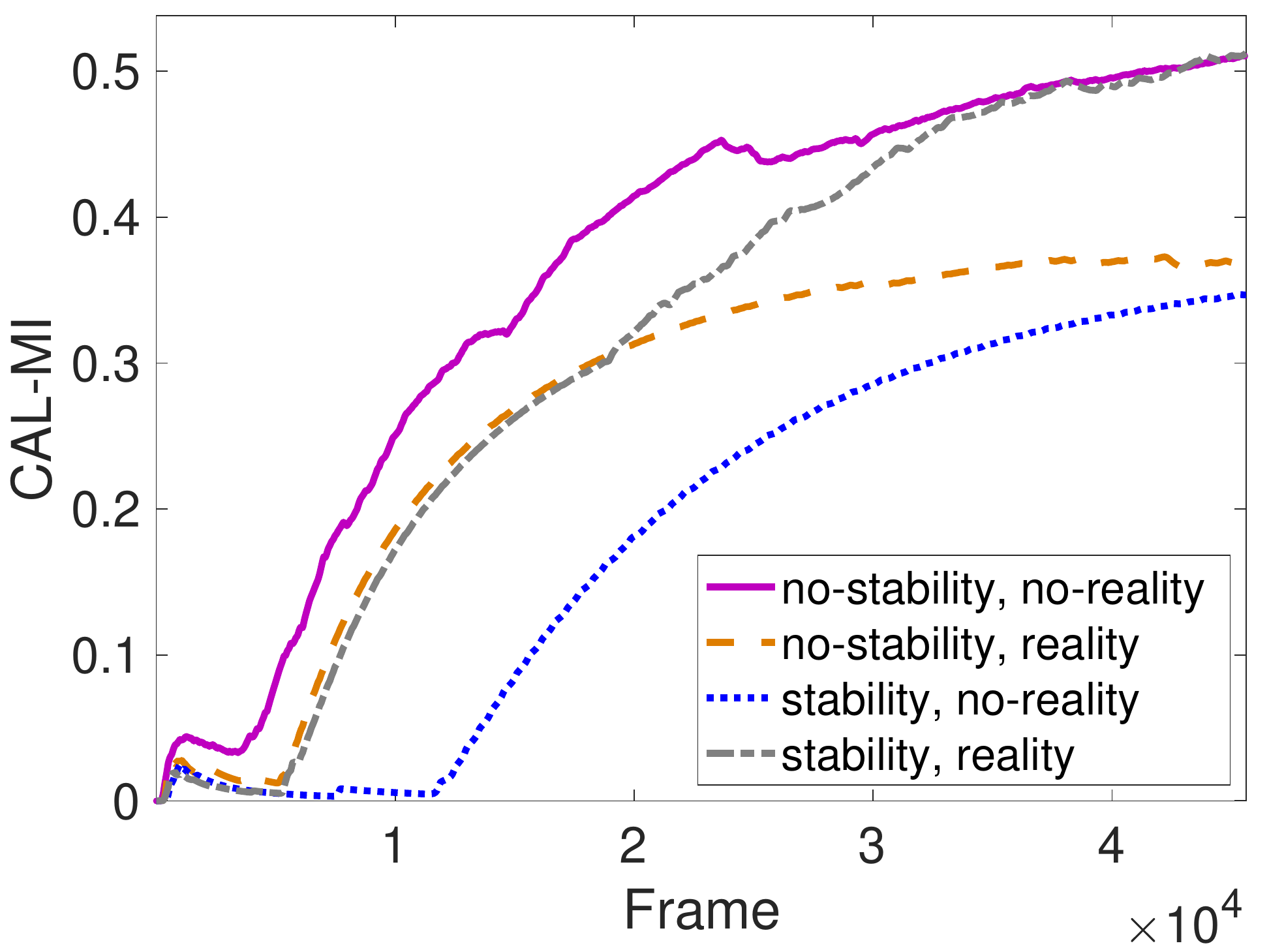>&\<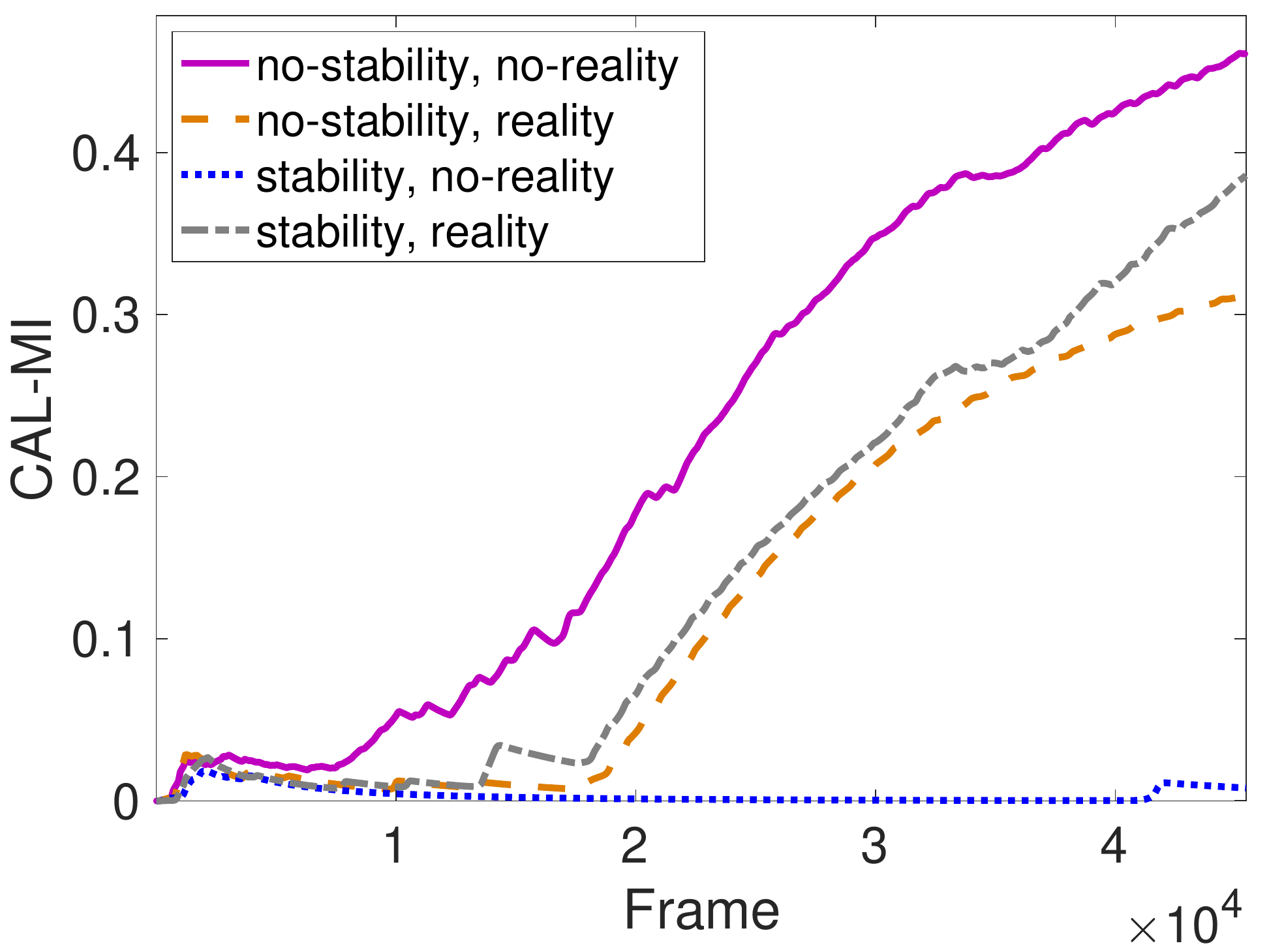>&
  \<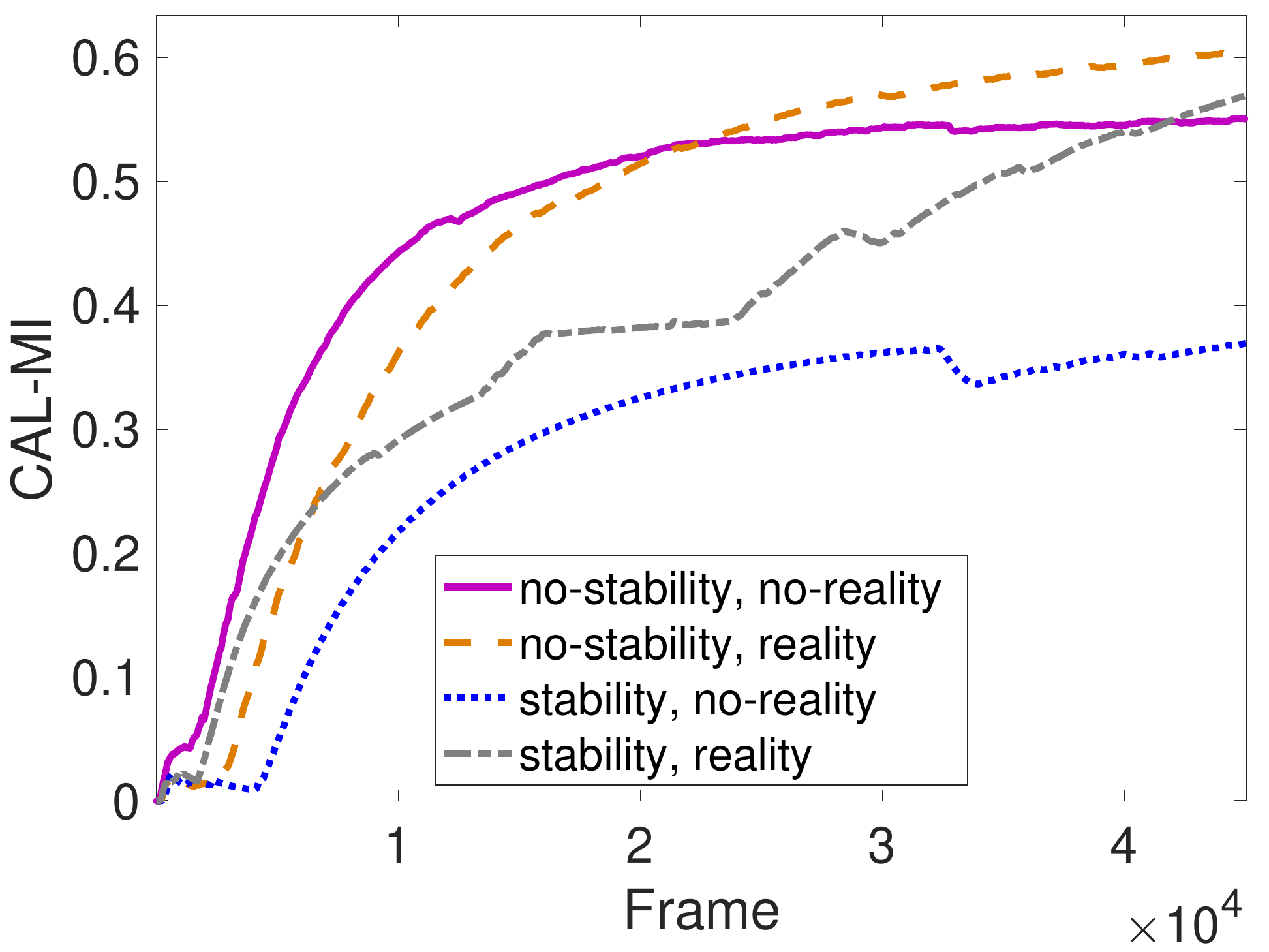>\cr
  \<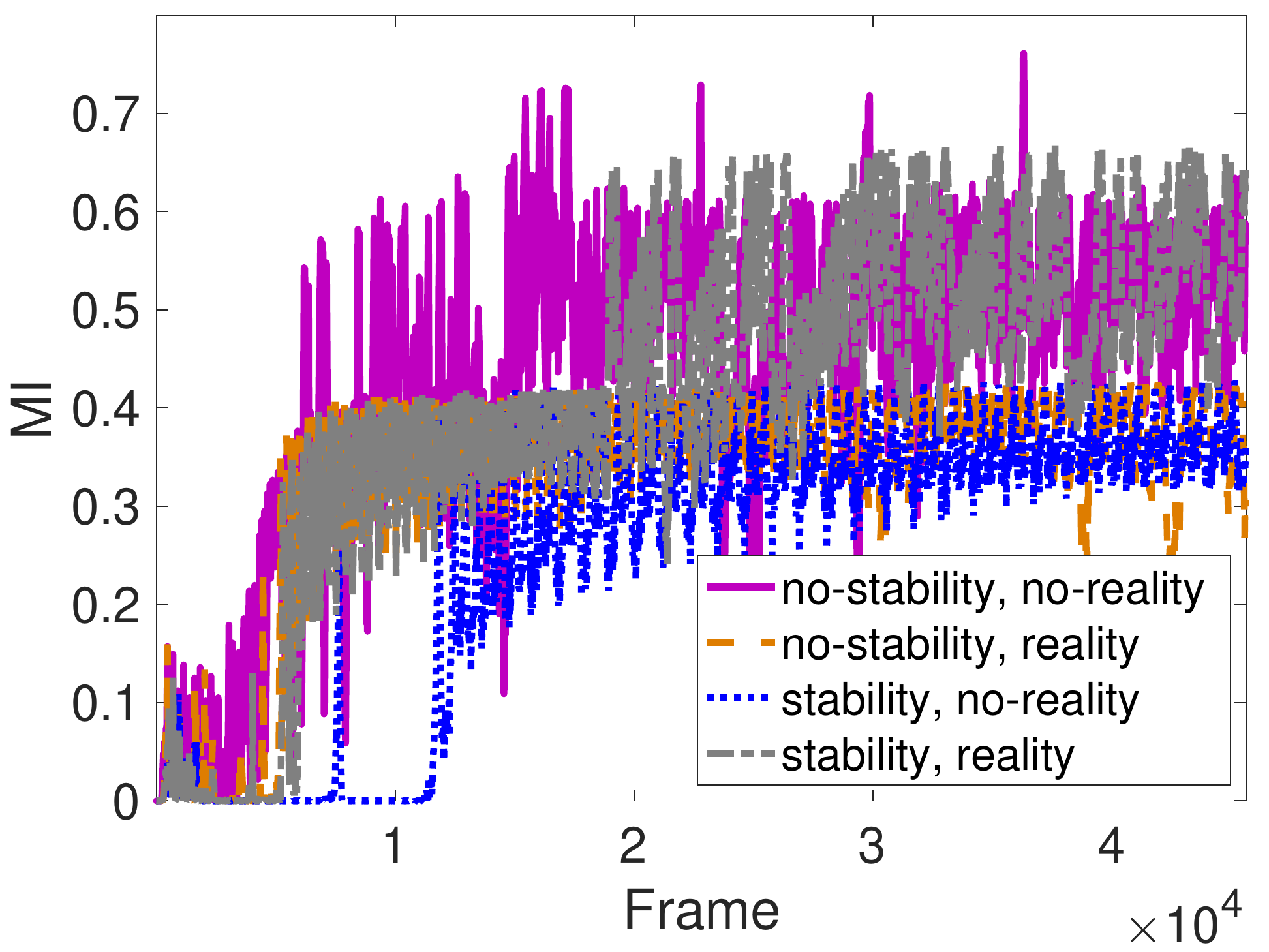>&\<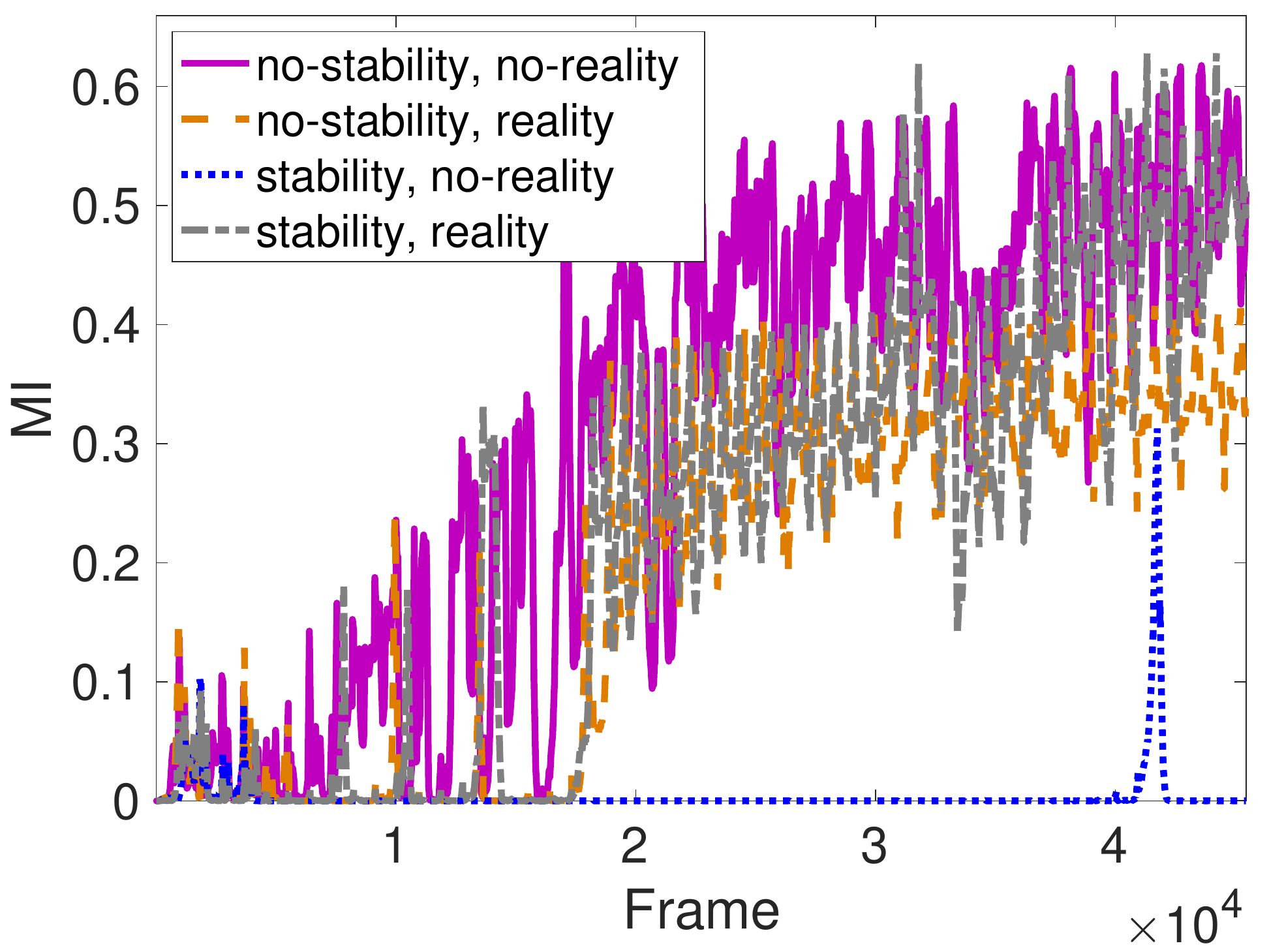>&
  \<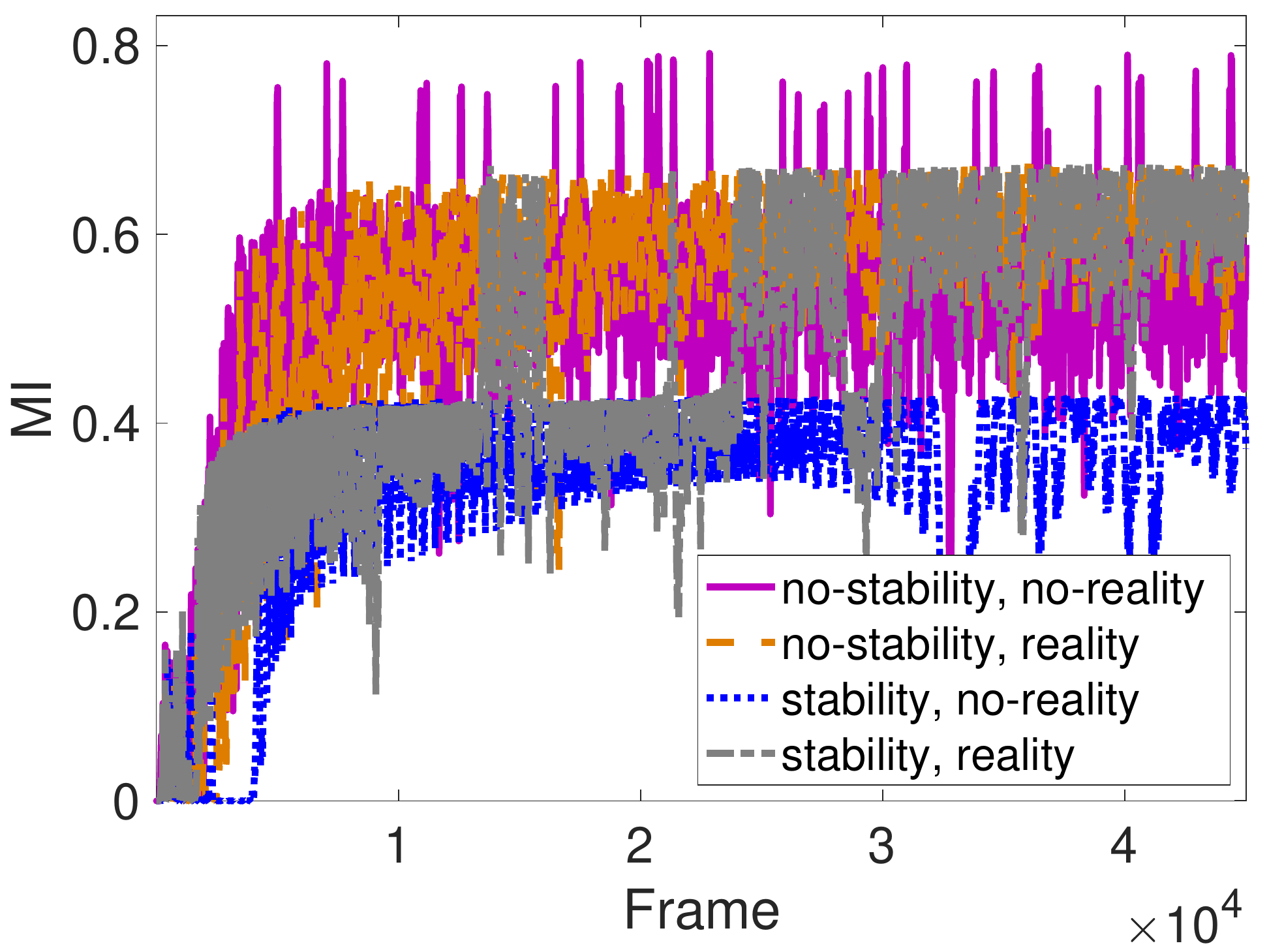>\cr
  \<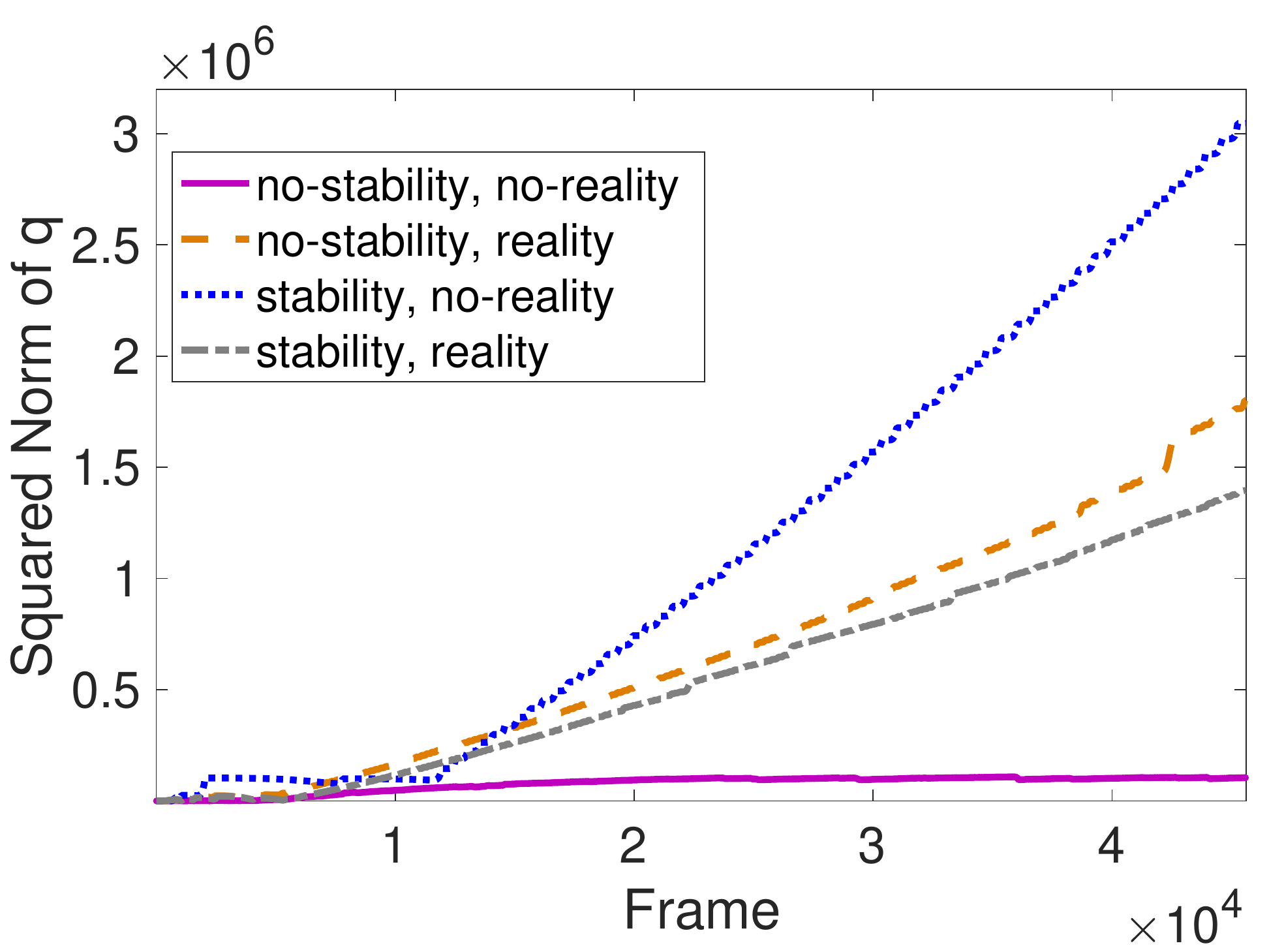>& 
  \<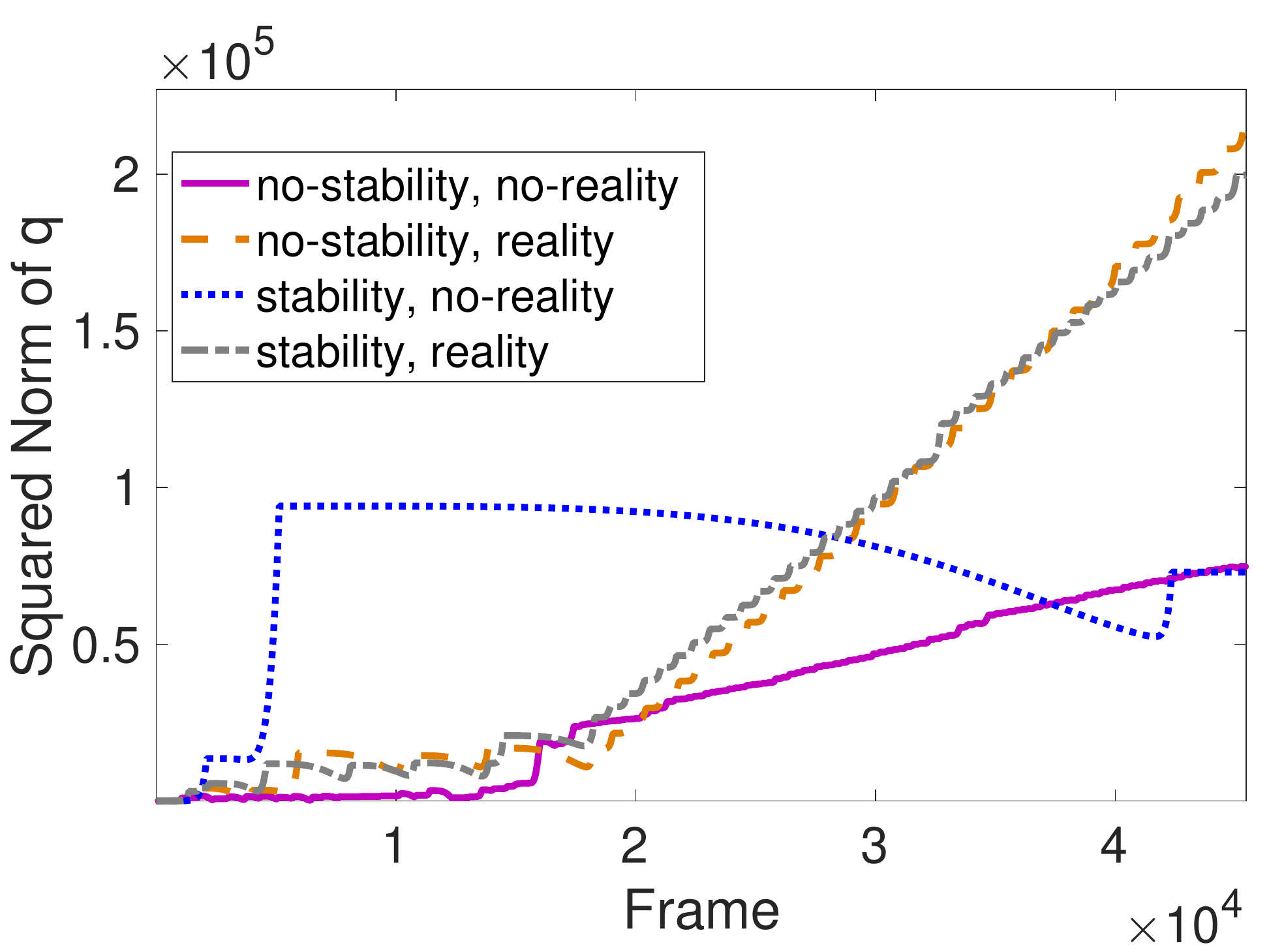>&\<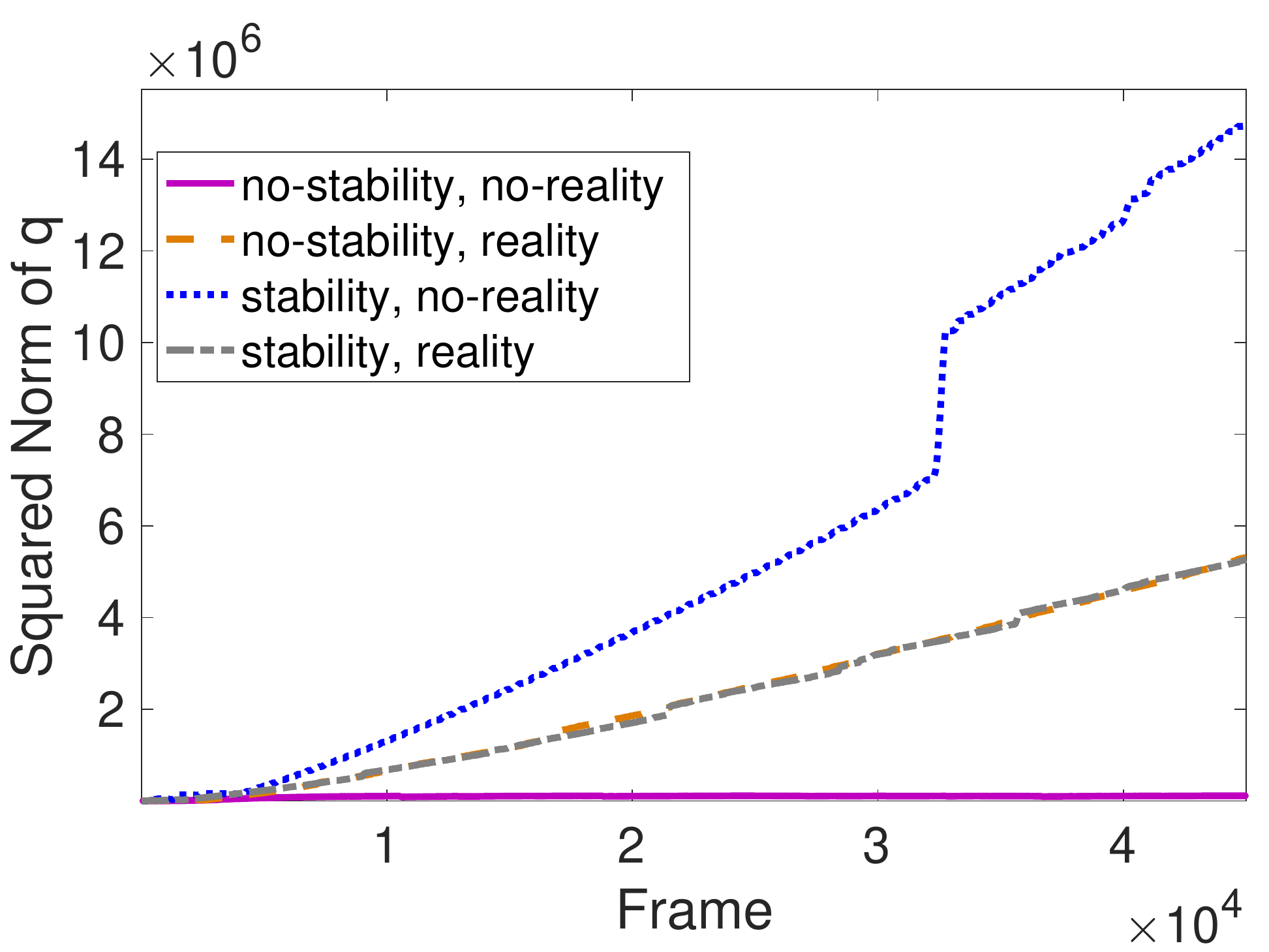>\cr
  \<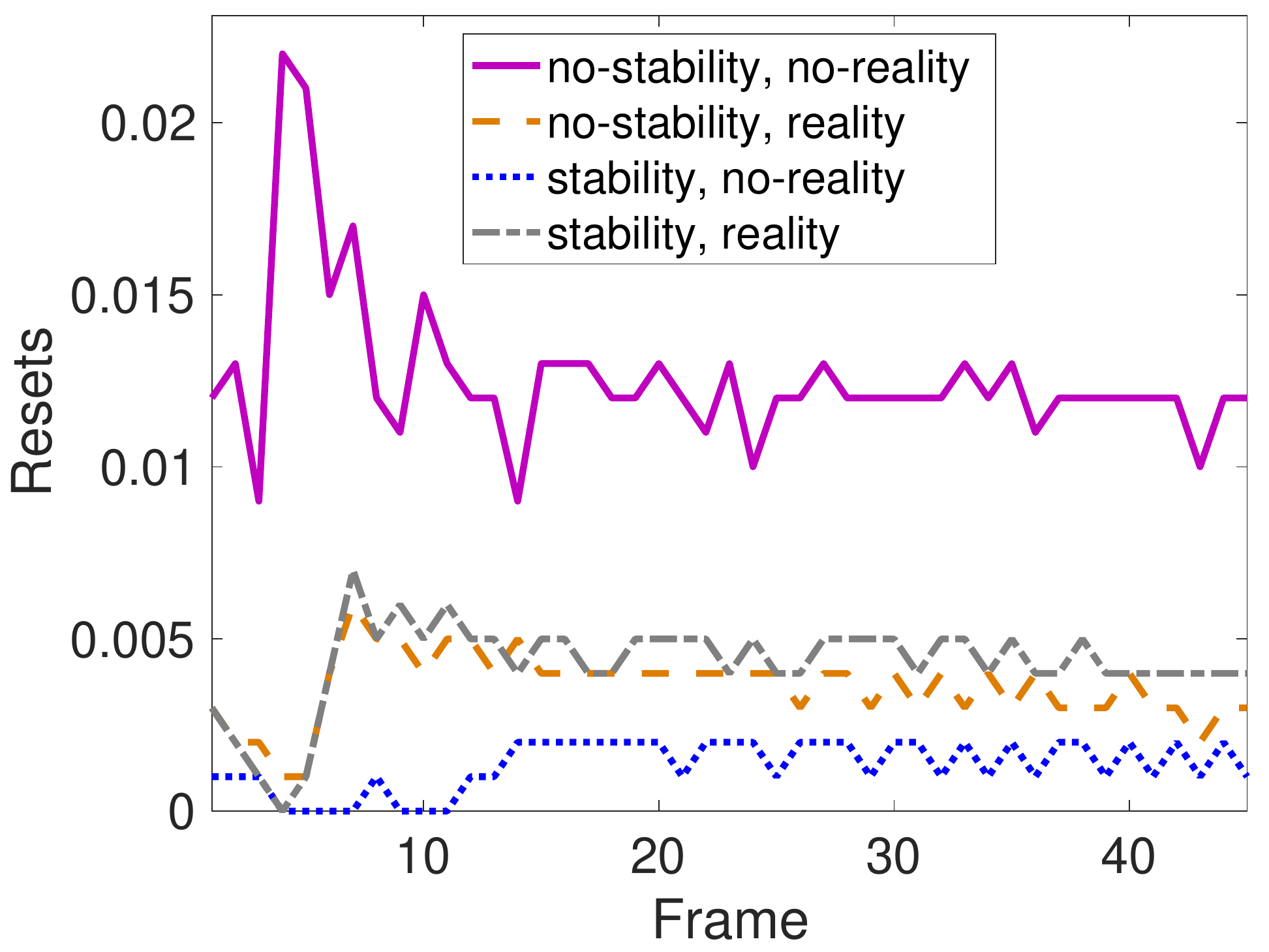>&
  \<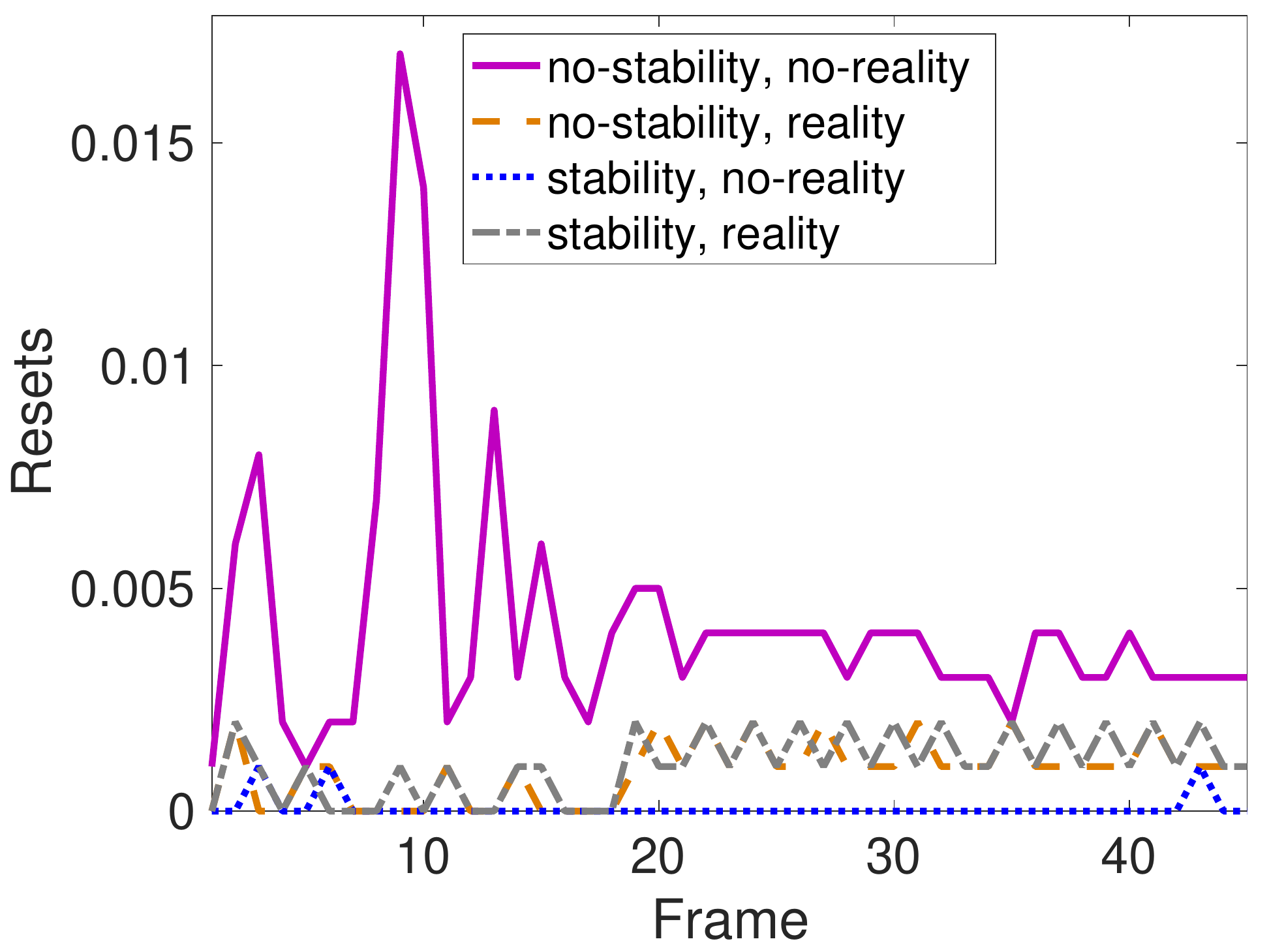>&
  \<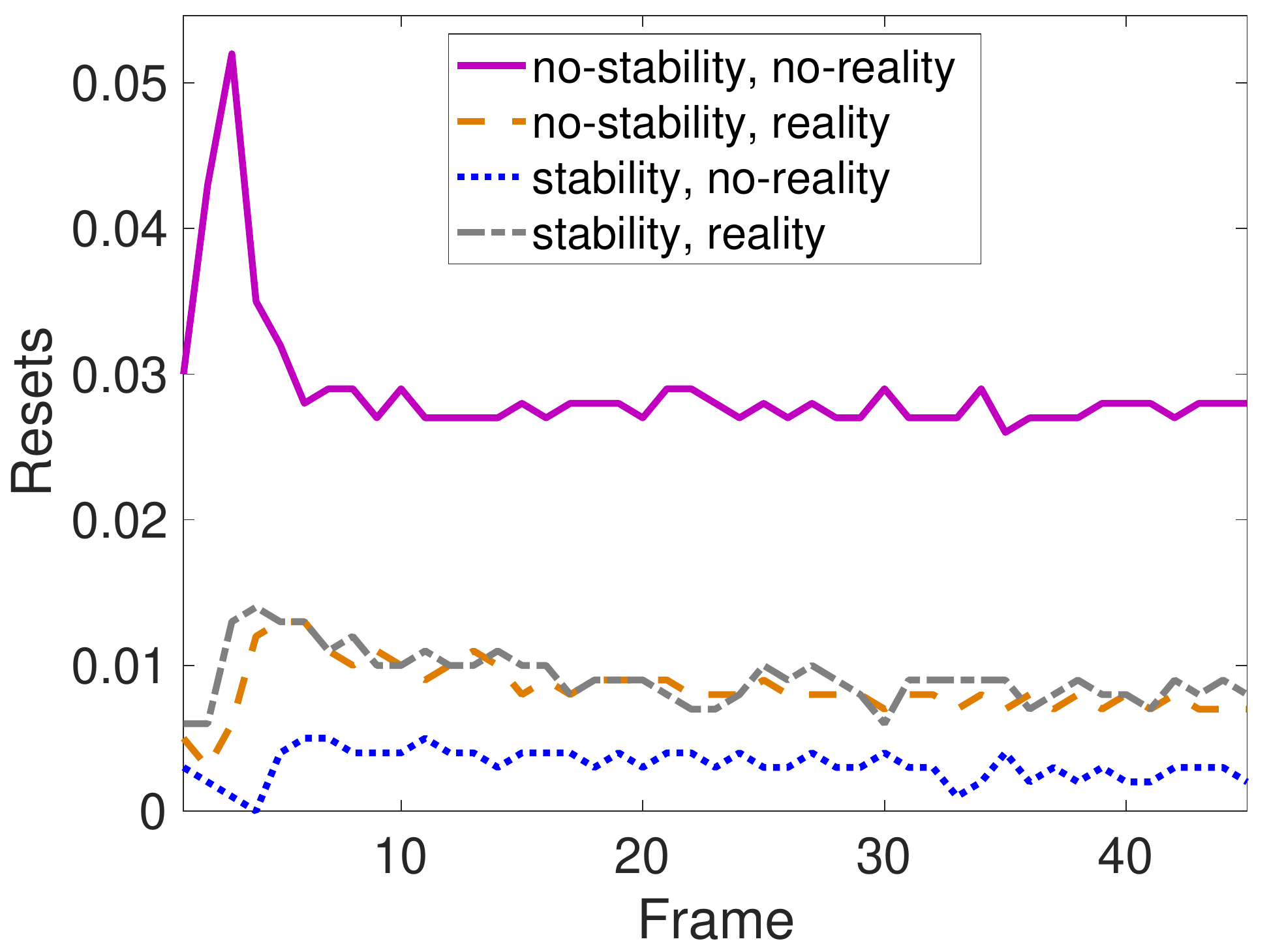>\cr
}
\caption{Comparing  4 configurations of the parameters, characterized by different properties in terms of stability and reality of the roots of the characteristic polynomial. The input video is reproduced (in loop) for 45k frames, at 3 different frame rates. The plots are organized into three columns, associated to the different frame rates. Each plot shows the temporal behavior measured by the frame index. From top-to-bottom, we report the cognitive action of 
Eq.~(\ref{cognitive-action-2}), the portion of the cognitive action that is about the Mutual Information (MI) (minus the potential), the MI per-frame, the norm of $q(t)$, and the fraction of ``reset'' operations performed every 1000 frames.}
\label{smallk} 
\end{figure*}

We performed experiments on the ``skater'' video clip; in particular we let
$h = 1 / \texttt{frame\_rate}$, $n=5$  features, 
and chose filters of size $5\times5$. Results are reported in Fig.~\ref{smallk} (first column). 
The plots indicate that there is an initial oscillation that is due to the effects of the blurring factor, that vanish 
after about 10k frames. The MI portion of the cognitive action correctly increases over time\footnote{When evaluating results, we used the classical MI based on the Shannon entropy.}, and it is pushed toward larger values in the two extreme cases of ``stability, reality'' and ``no-stability, no-reality''. The latter shows more evident oscillations in the frame-by-frame MI value, due to the not-stable configuration, and roots with imaginary part. In all the configurations the norm of $q$ increases over time, due to the small values of $k$, while the frequency of reset operations is larger in the ``no-stability, no-reality'' case. When moving to the second column of Fig.~\ref{smallk} (smaller frame rate), we can see that all the configurations have more difficulties in reducing the cognitive action and increasing the MI index. This is due to the faster changes in the video signal and the larger $h$ that makes it hard to follow the dynamics of the CAL (leading to a bad solution in the case of ``stability, no-reality''). Differently, when using a larger frame rate (third column of Fig.~\ref{smallk}), we get better results, that seem to support our intuition of slowly presenting information to the system. The system is also able to develop the MI index in a smaller number of steps. 

\begin{figure*}
  \hbox to\hsize{\hfil\<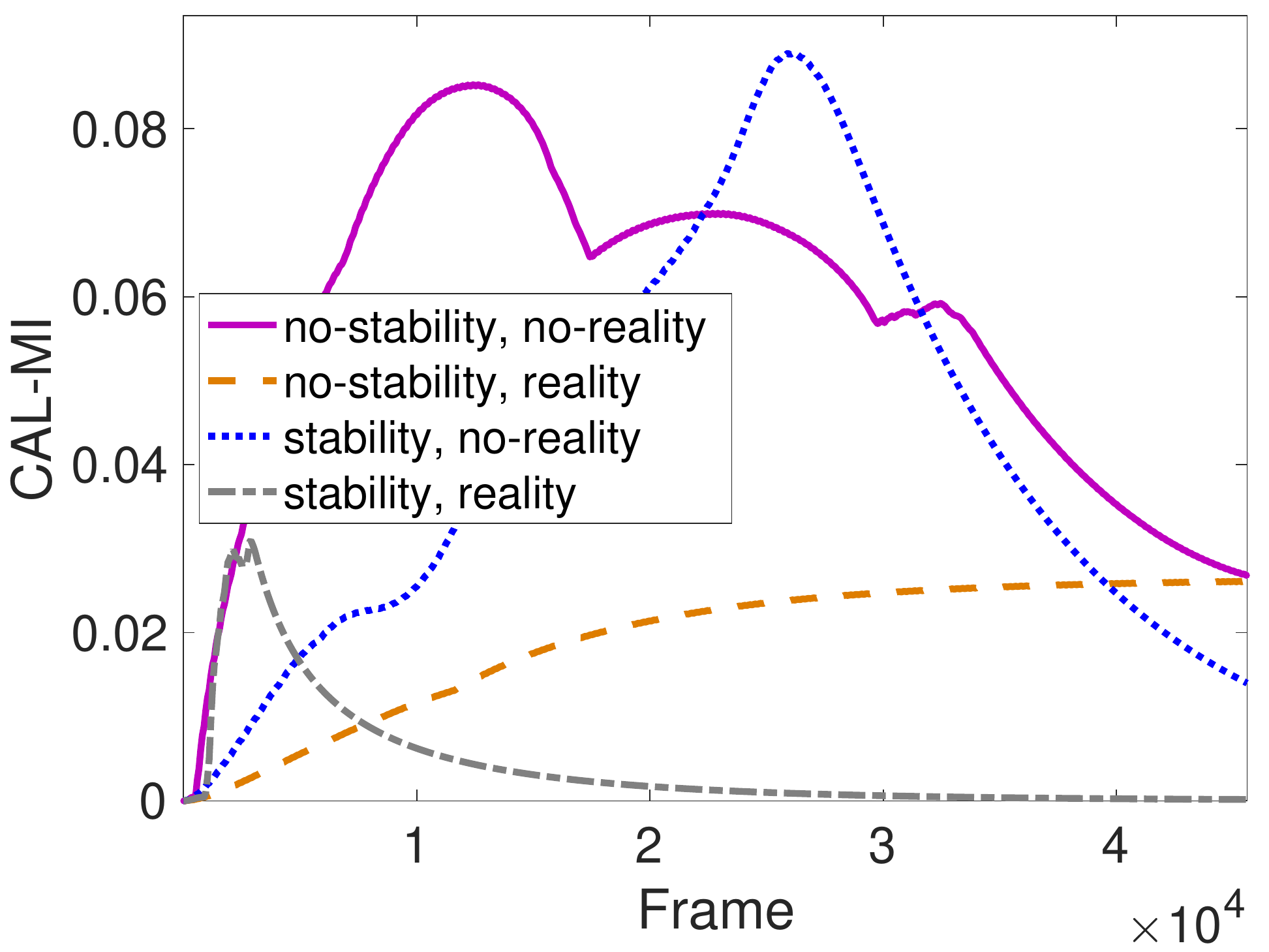>\hfil\<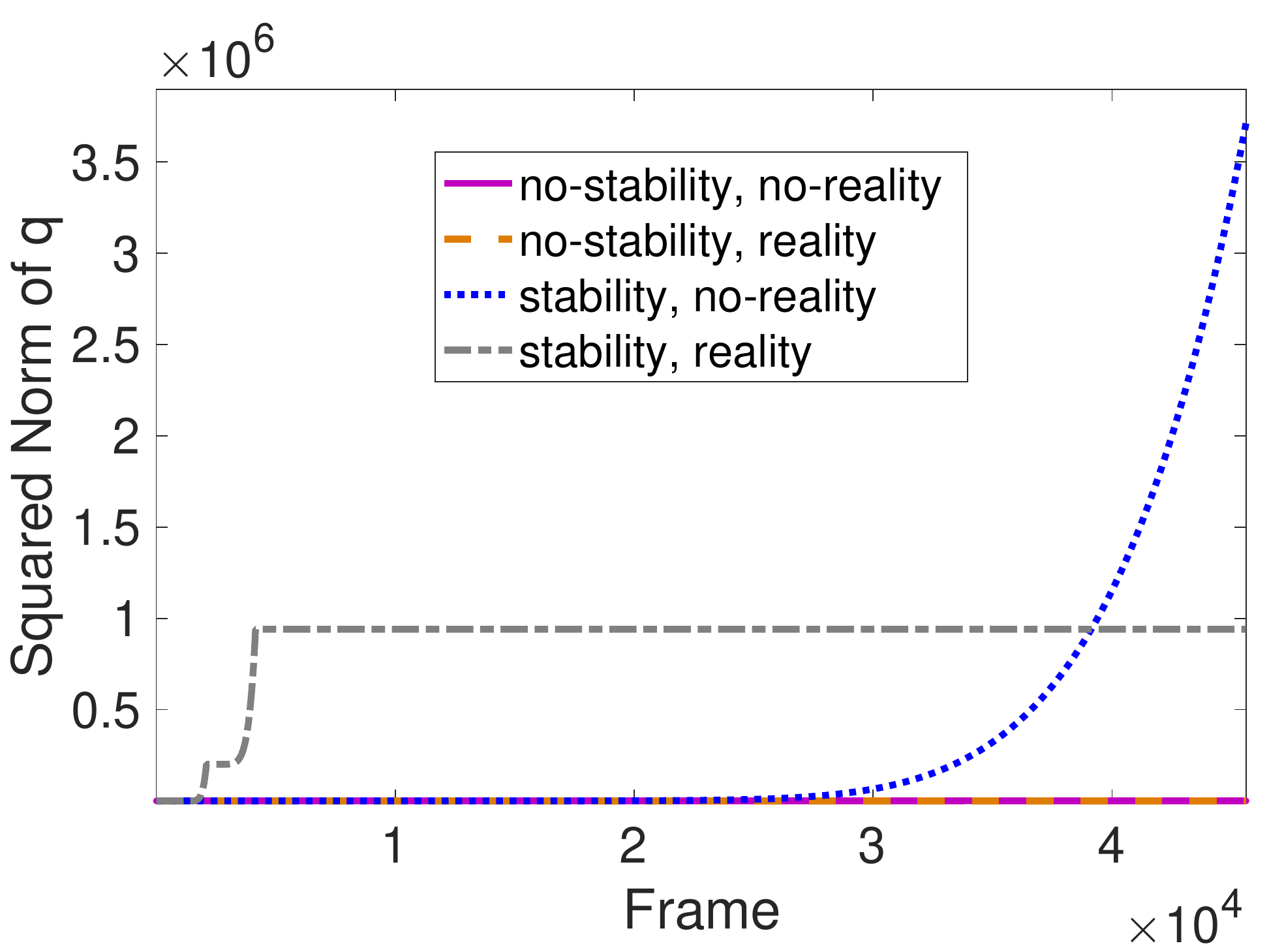>\hfil
    \<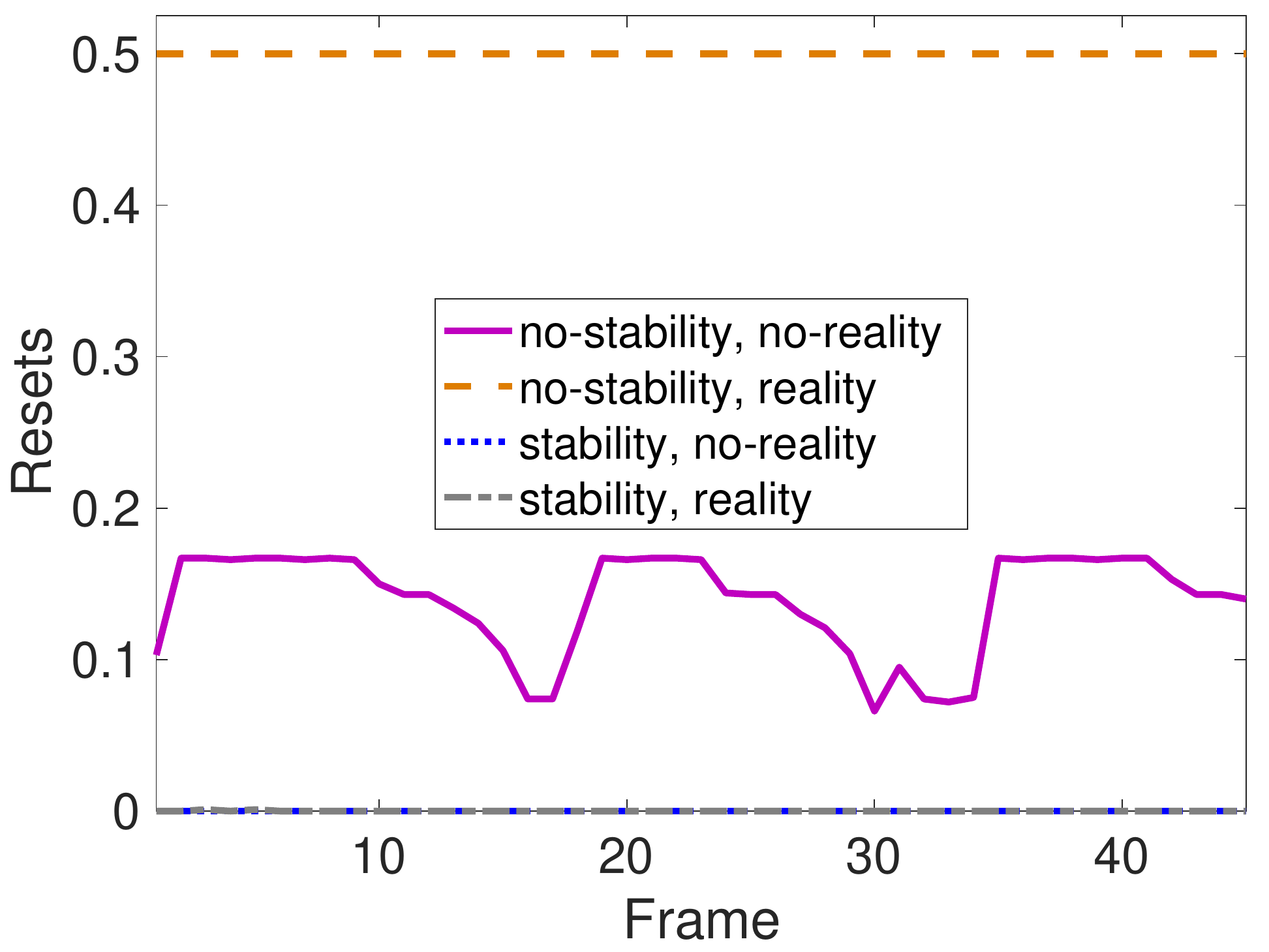>\hfil}
\vskip -3mm
\caption{Comparing 4 new configurations of the parameters, all of them with ``larger'' values of $k$ (with respect to the ones of Fig.~\ref{smallk}).}
\label{largek}
\end{figure*}

We investigated other configurations of parameters that are characterized by larger values of $k$ (between 10 and 20, in the not-stable configurations, and of the order of $10^{-8}$ and $10^{-15}$ in the last two configurations, respectively). Fig.~\ref{largek} shows that the MI index is always pretty small. This is due to the stronger regularization that we enforce in the problem, so that the system has difficulties in developing good features.
However, the norm of $q$ is either small or it becomes almost constant after awhile (with the exception of one configuration, where $q$ still grows), showing the convergence of the variables to a fixed value. We clearly observe that the unstable configurations make a wider use of the reset mechanism.
We evaluated the quality of the developed features by freezing the final $q$ of Fig.~\ref{smallk} and computing the MI index over a single repetition of the whole video clip, reporting the results in Tab.~\ref{mivw} (a). We notice that, while in Fig.~\ref{smallk} we compute the MI on a frame-by-frame basis, here we compute it over the whole frames of the video at once, thus in a batch-mode setting. The result confirms that the 100 fps case is preferable, and that the two extreme configurations ``stability, reality'' and ``no-stability, no-reality'' show better results, on average. While this was expected in the ``stability, reality'' case, we explain the  performances of ``no-stability, no-reality'' by the effect of the reset mechanism, that allows even such unstable configuration to develop good solutions.

\begin{figure*}
  \tabskip=1em plus2em minus.5em
  \halign to\hsize{\hfil#\hfil&\hfil#\hfil&\hfil#\hfil\cr
    \<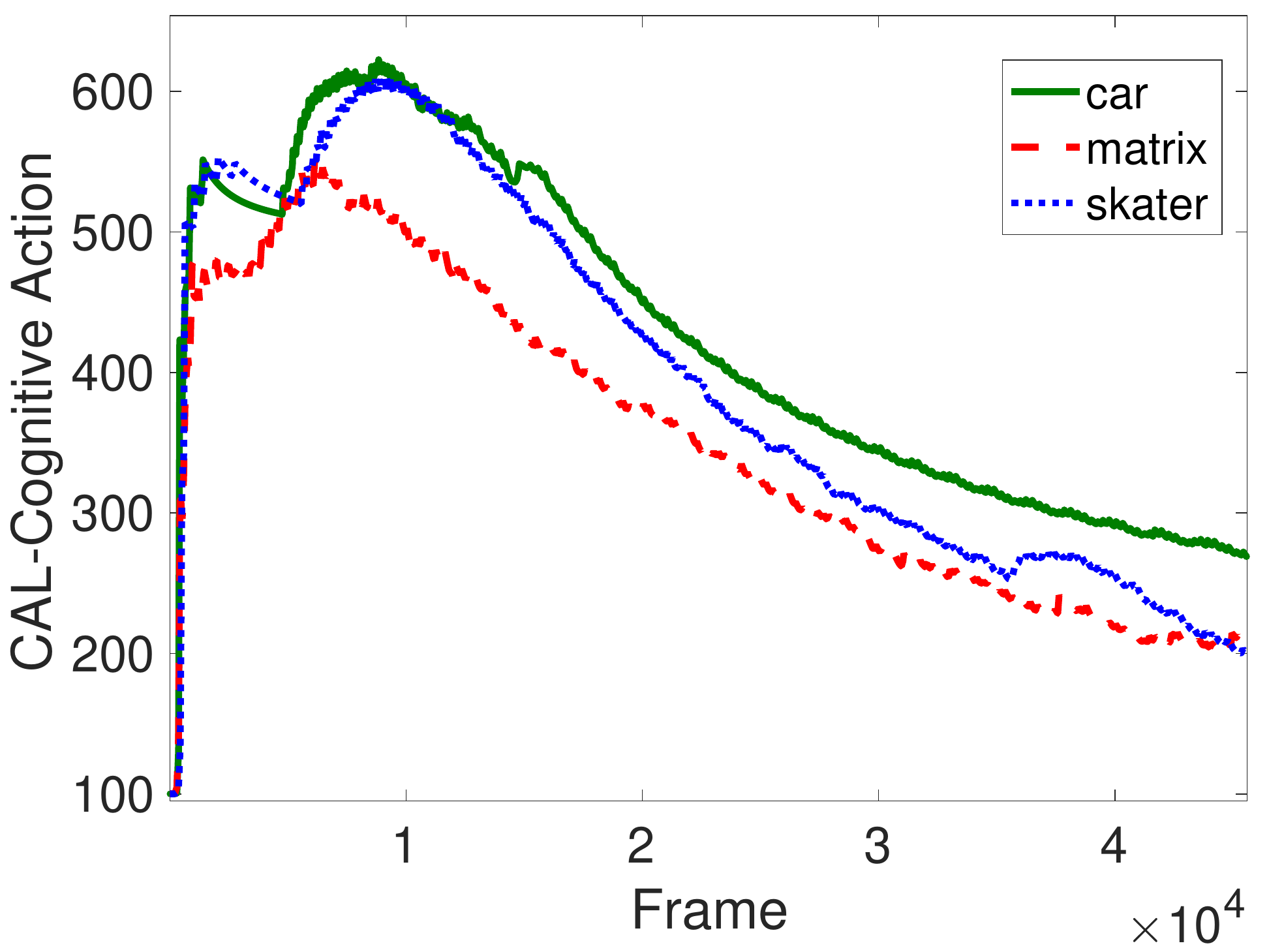>&\<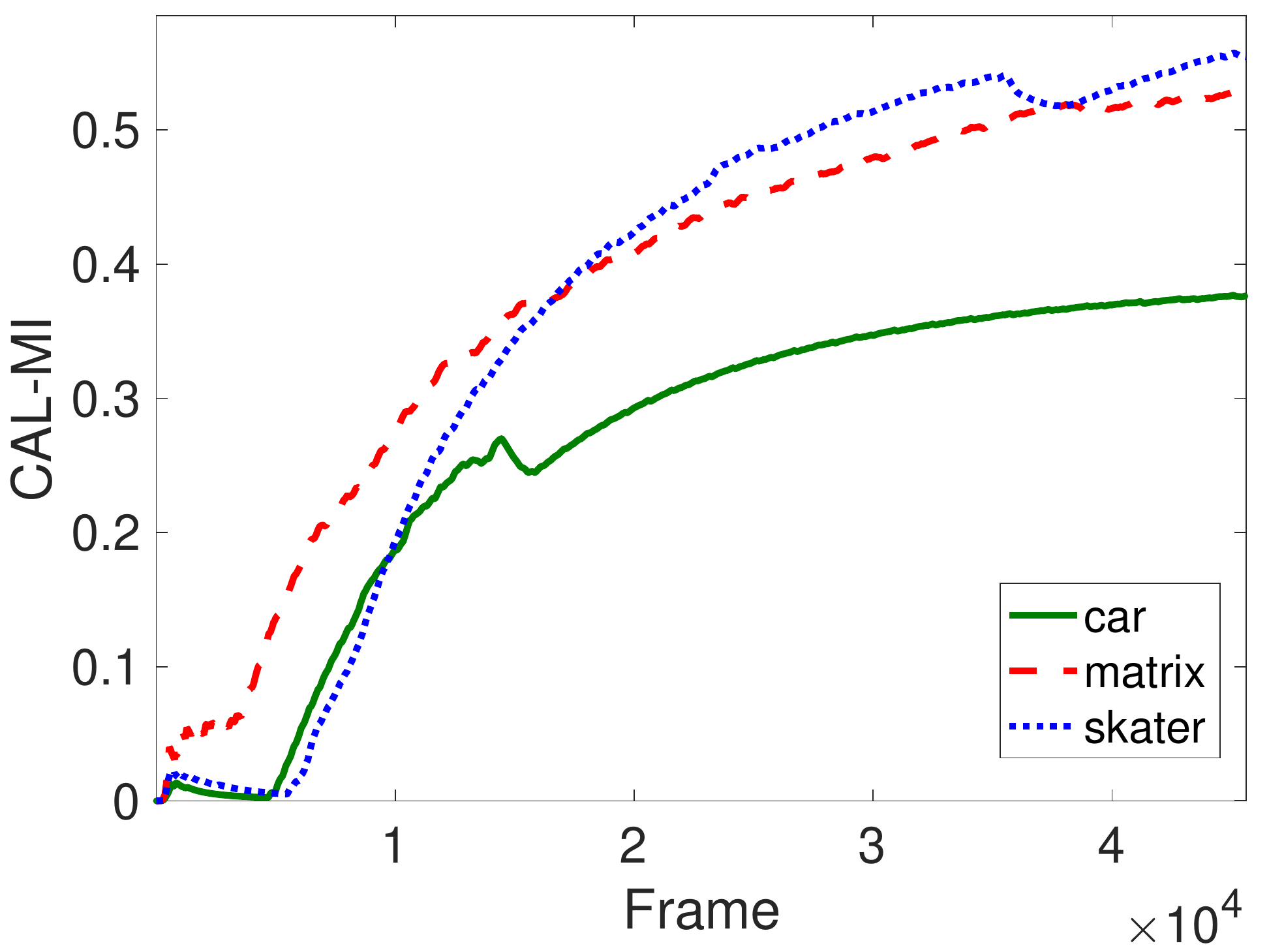>&\<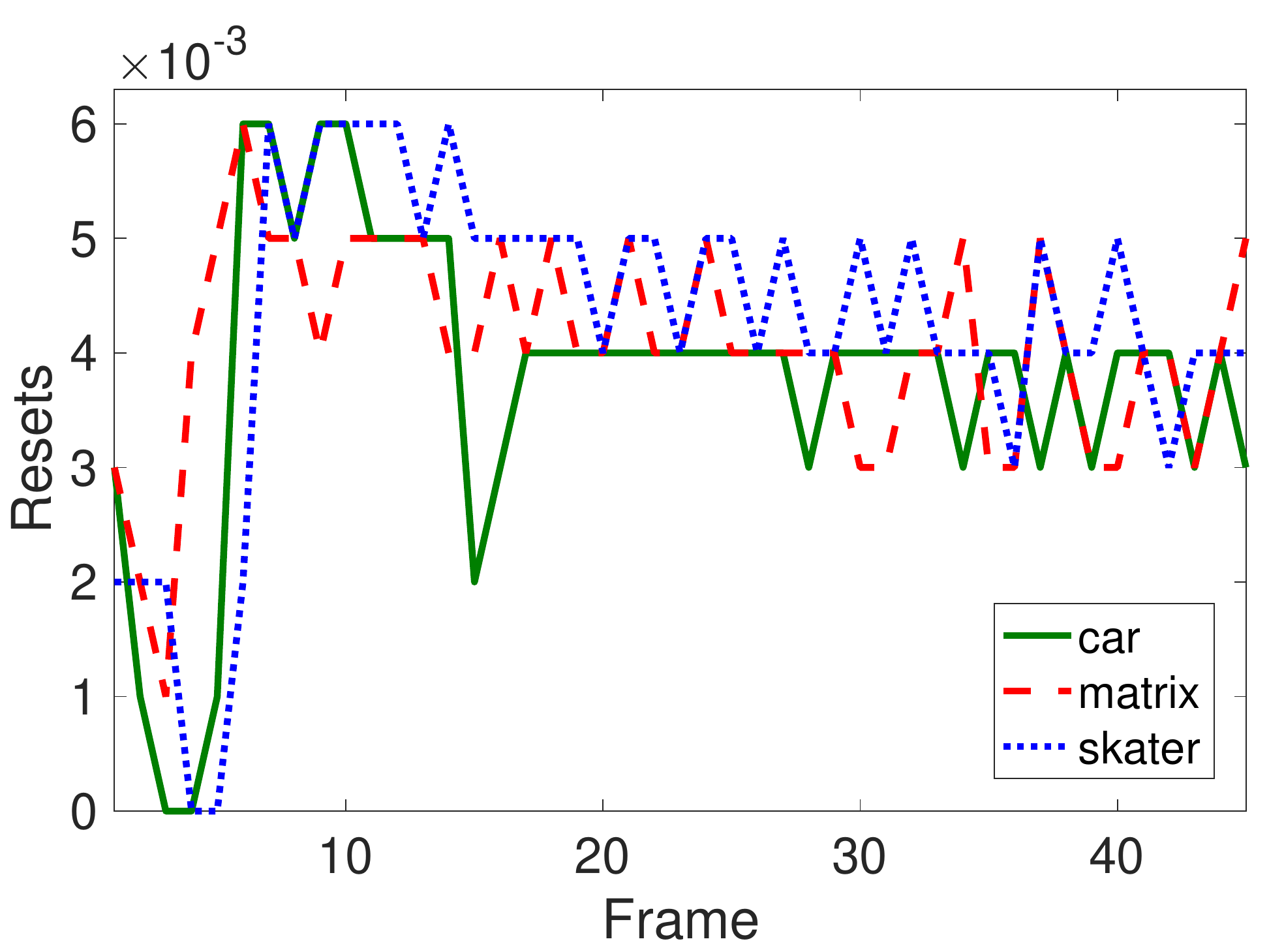>\cr
    \<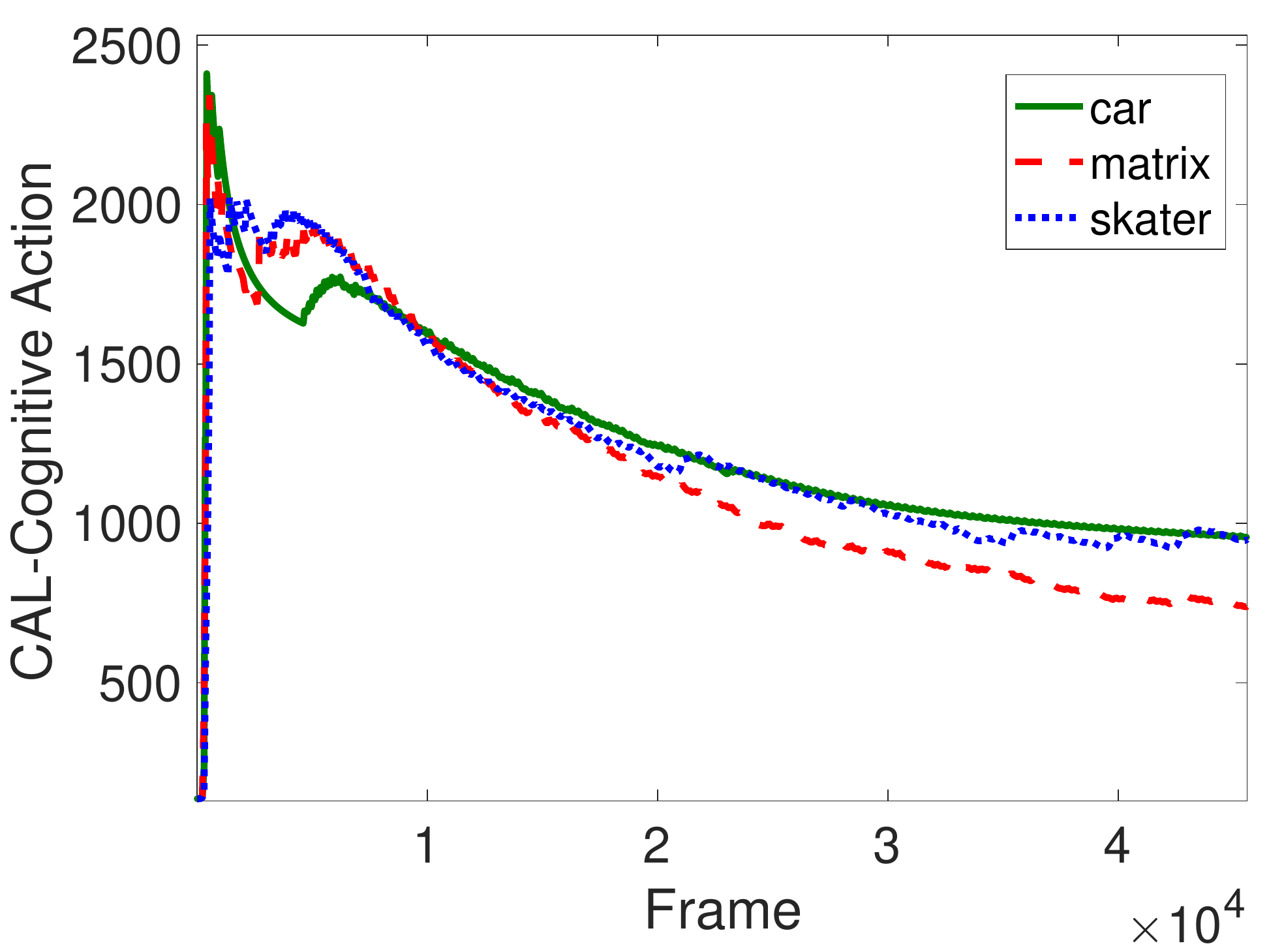>&\<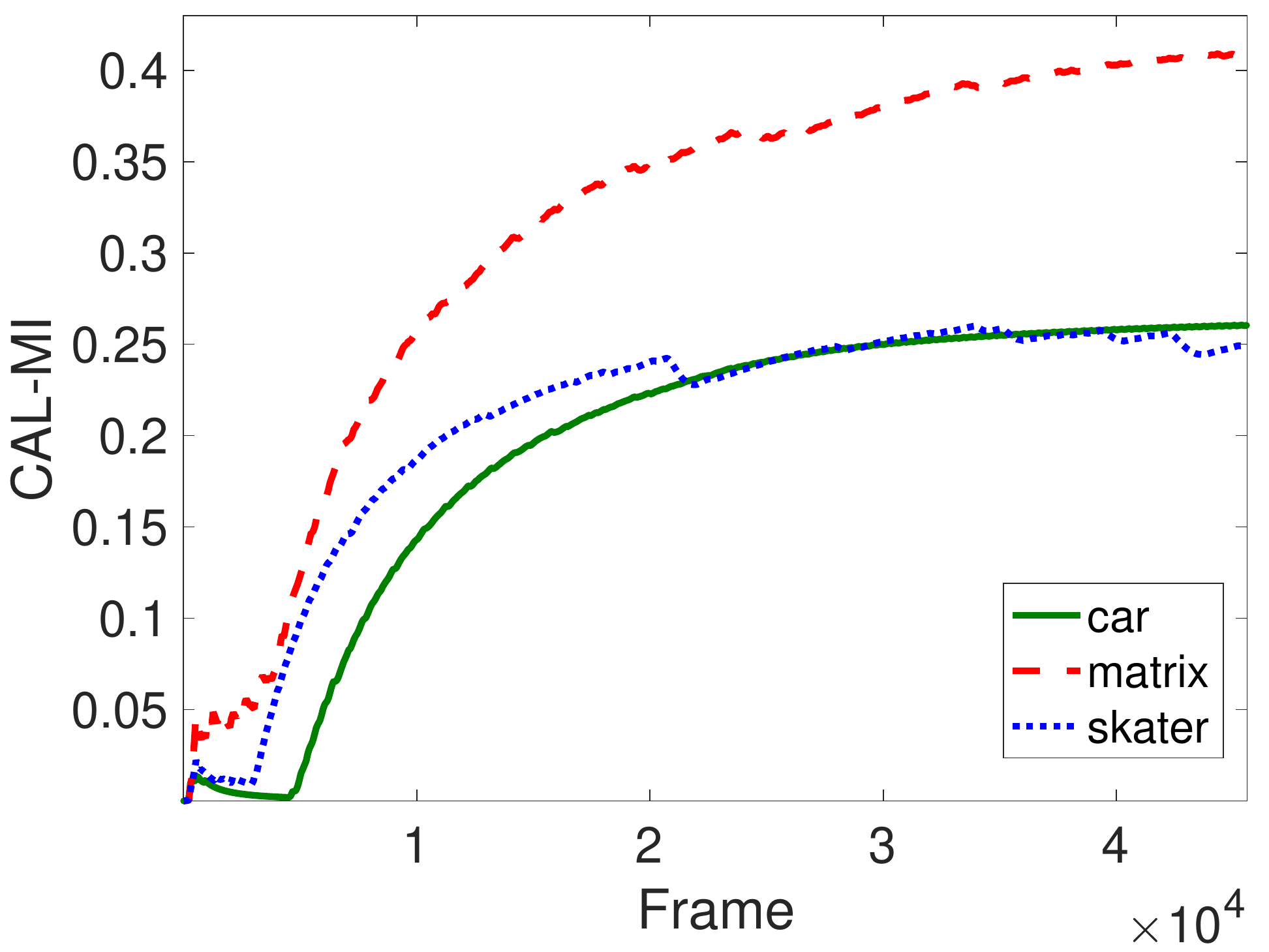>&\<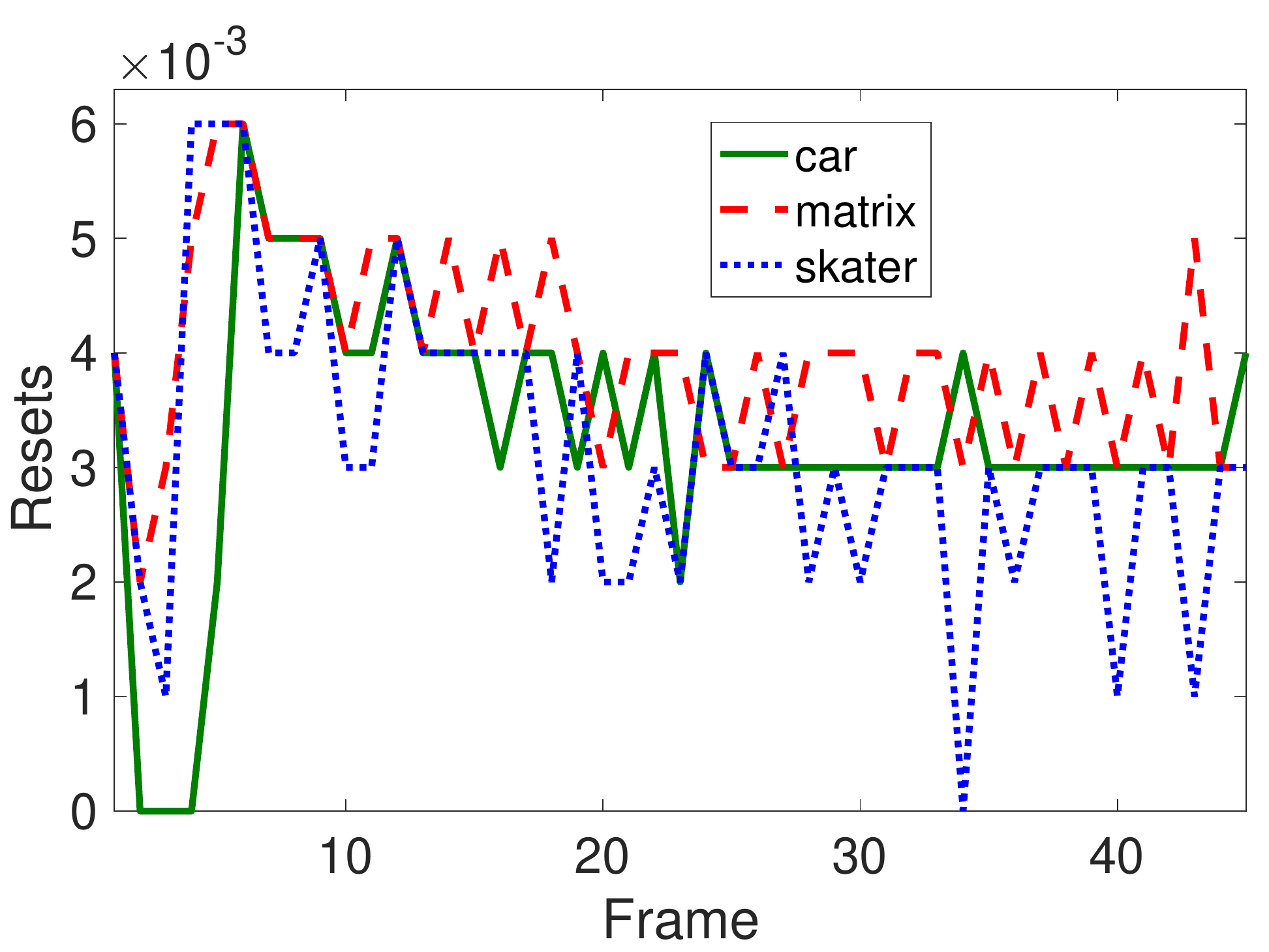>\cr}
\vskip -3mm
\caption{Different number of features and filter sizes (1st row: $n=5, size=5\times5$; 2nd row: $n=11, size=11\times11$) in 3 videos. See Fig. \ref{smallk} for a description of the plots.}
\label{vids}
\end{figure*}

\begin{figure}
    \hbox to\hsize{\hfil\|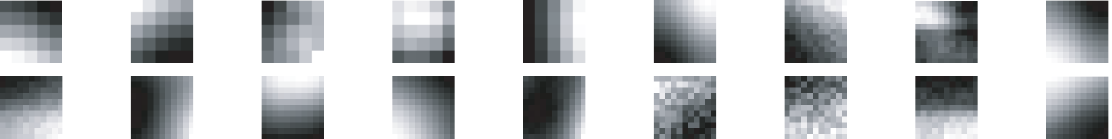| \hfil}
\vskip -1mm    
\caption{Some pictures of the filters obtained by solving CAL. The first five filters (top row) are from the $5\times5$ case, the others are $11\times11$.}
\label{demo}
\end{figure}

We compared the behavior of the system on multiple video clips and using different filter sizes ($5\times5$ and $11\times11$) and number of features ($n=5$ and $n=11$) in Fig~\ref{vids}. We selected the ``stability, reality'' configuration of Fig.~\ref{smallk}, that fulfils Proposition \ref{prop:coef}. Changing the video clip does not change the considerations we did so far, while increasing the filter size and number of features can lead to smaller MI index values, mostly due to the need of a better balancing the two entropy terms ($\lambda_E$) to cope with the larger number of features. The MI of Table~\ref{mivw} (b) confirms this point. Interestingly, the best results are obtained in the longer video clip (``The Matrix'') that requires less repetitions of the video, being closer to the real online setting. In Fig.~\ref{demo} we repot some of the developed filters, that clearly resemble oriented edges and corners.

\begin{figure*}
  \hbox to\hsize{\hfil\<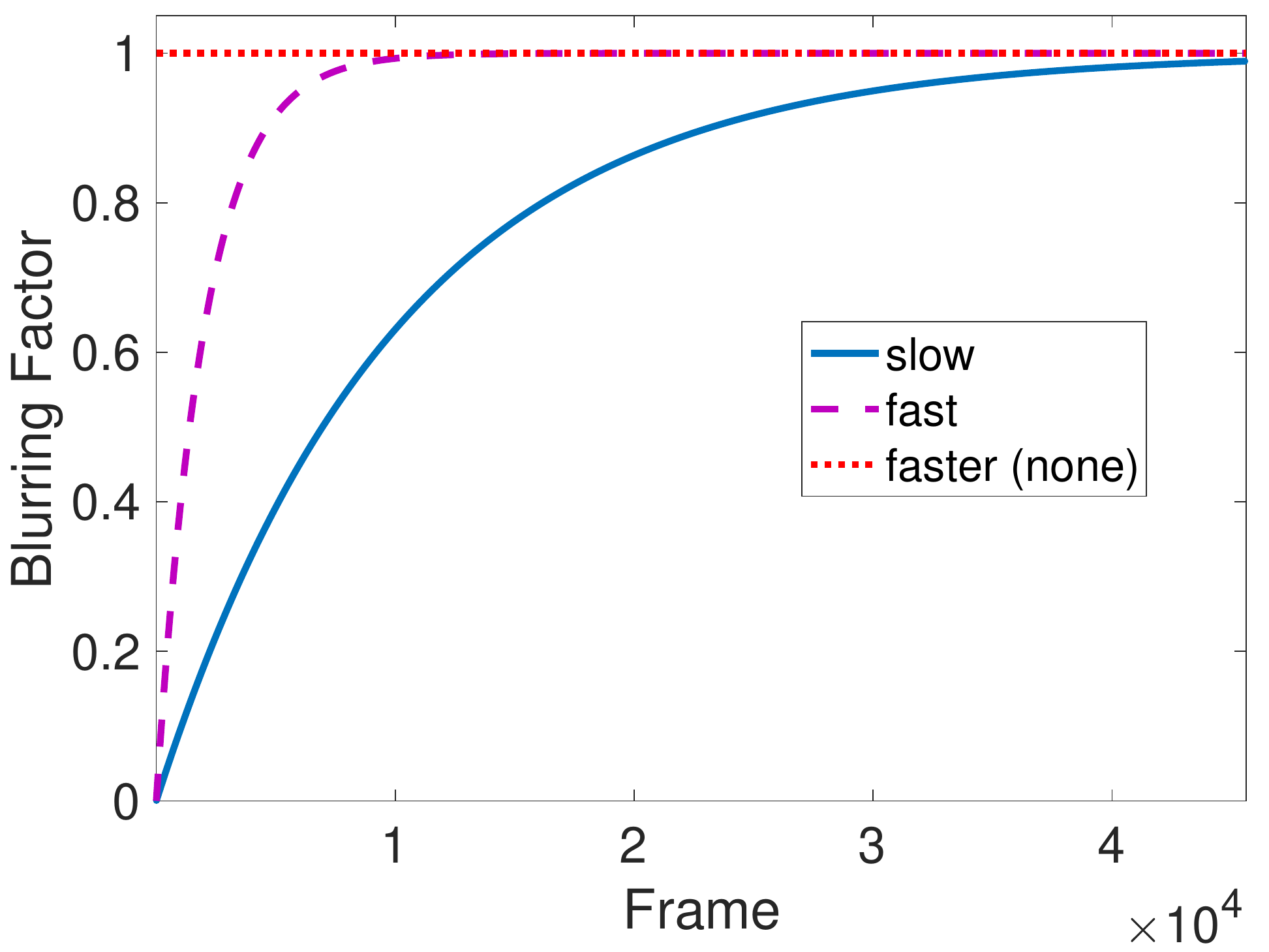>\hfil\<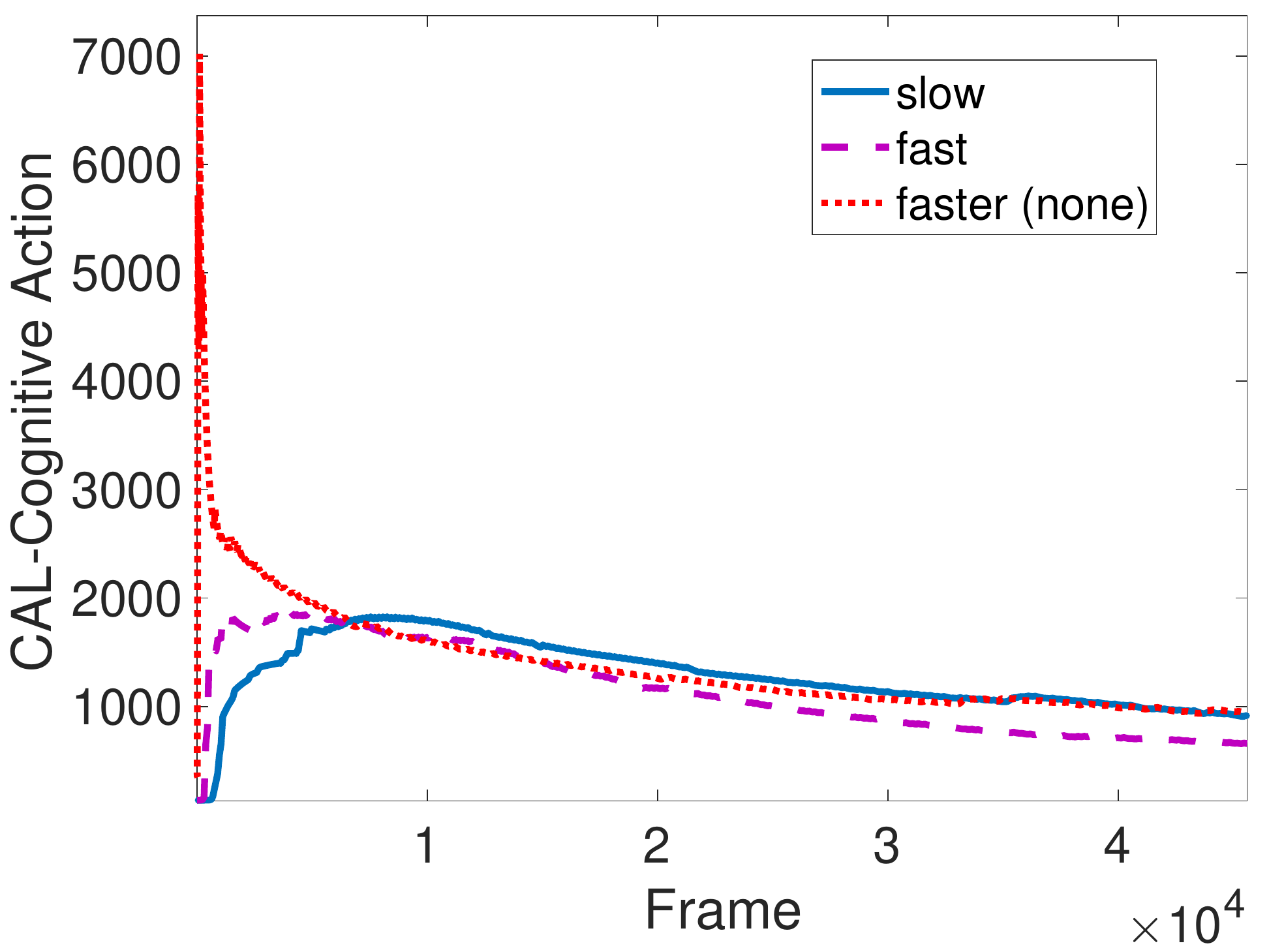>\hfil
    \<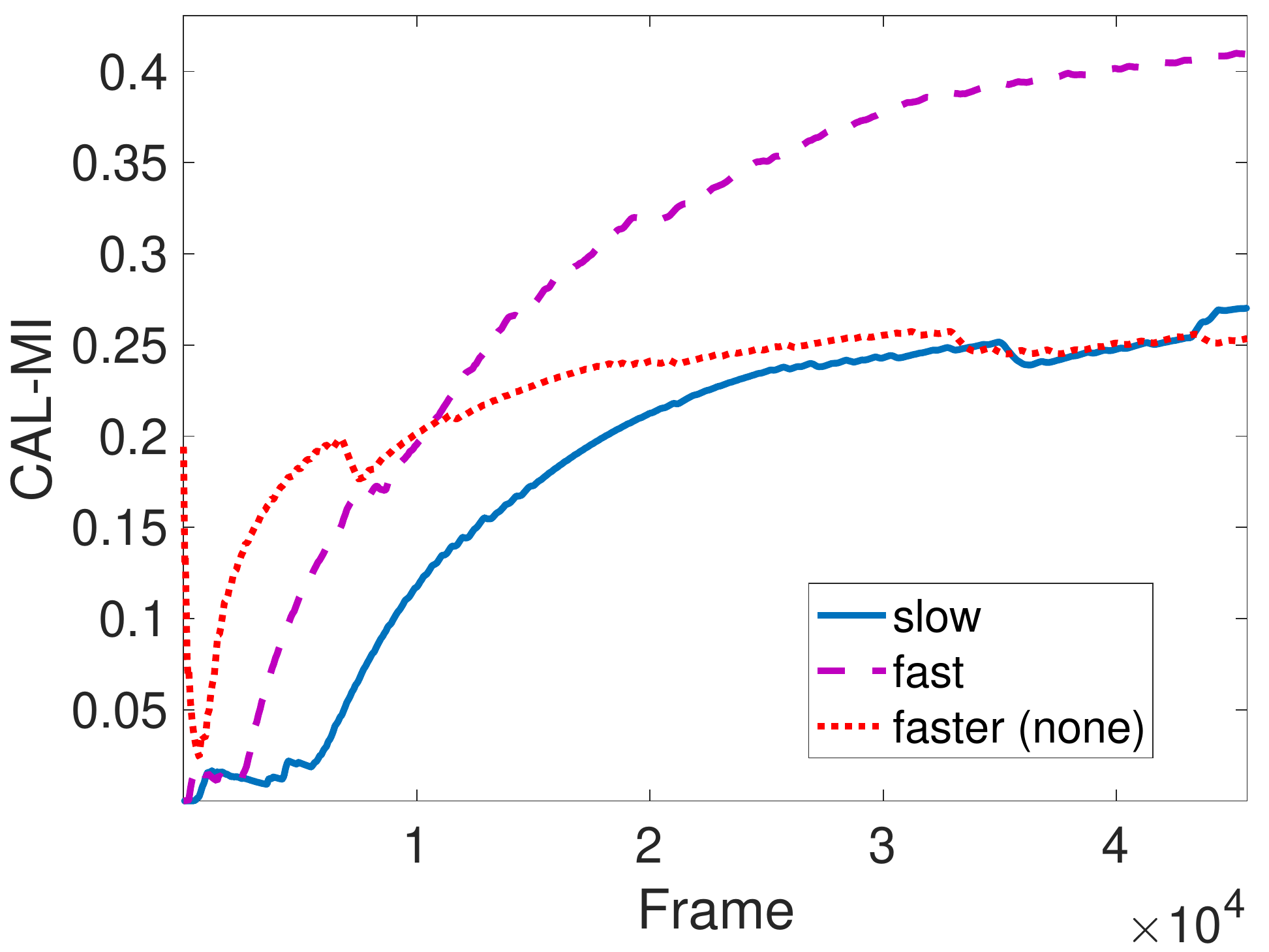>\hfil}
\vskip -3mm
\caption{Three different developmental plans ($n=11$ and filters of size $11\times11$).}
\label{blurs}
\end{figure*}

Fig.~\ref{blurs} and Table \ref{mivw} (c) show the results we obtain when using different developmental plans (``skater'' clip), that is, different values of $\eta$ that lead to the blurring factors reported in the first graph of Fig.~\ref{blurs}. These results suggest that a gradual introduction of the video signal helps the system to find better solutions than in the case in which no-plans are used, but also that a too-slow plan is not beneficial. The cognitive action has a big bump when no-plans are used, while this effect is more controlled and reduced in the case of both the slow and fast plans.

\begin{figure}
\hbox to\hsize{\hfil\|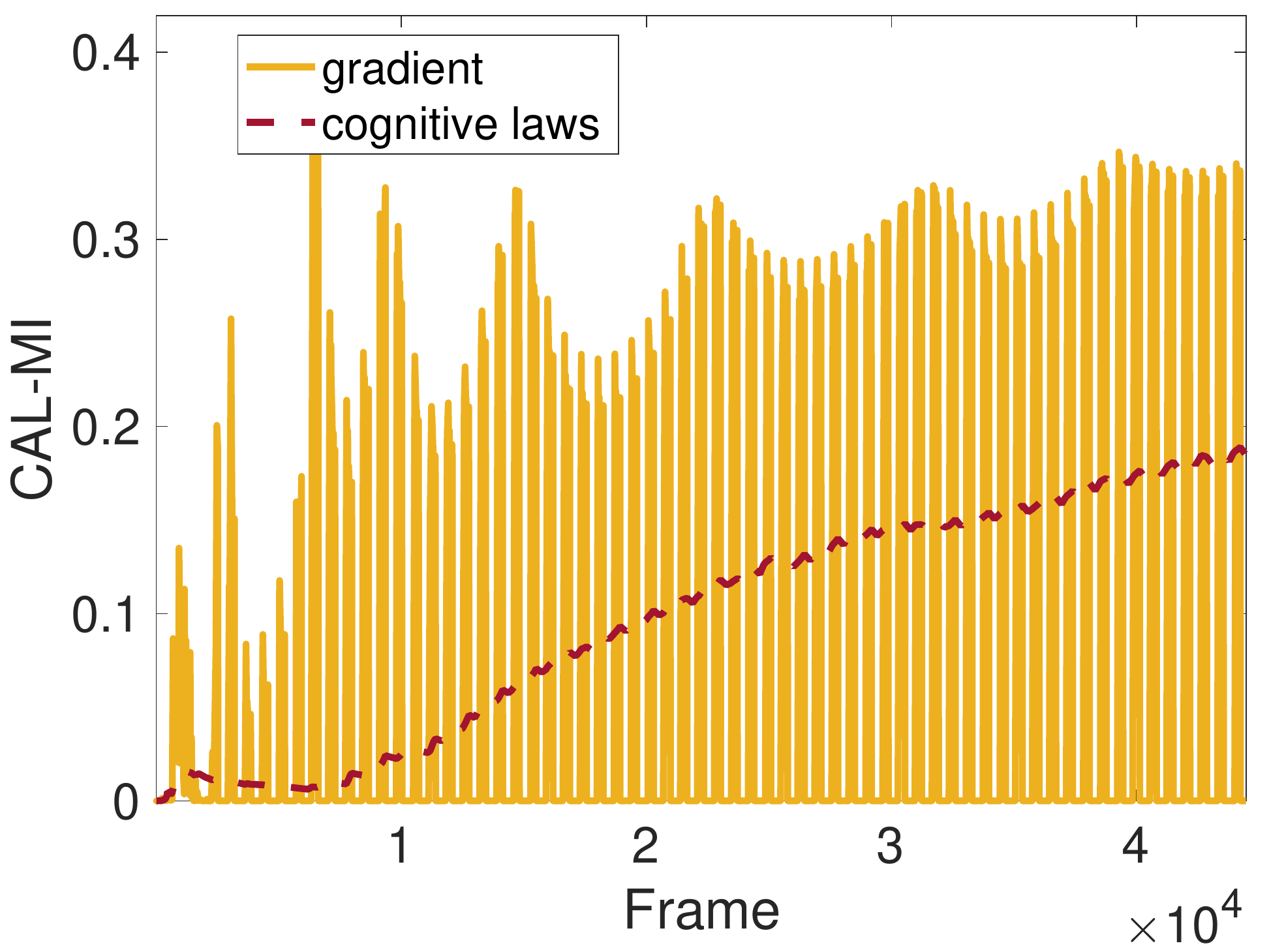|\hfil\|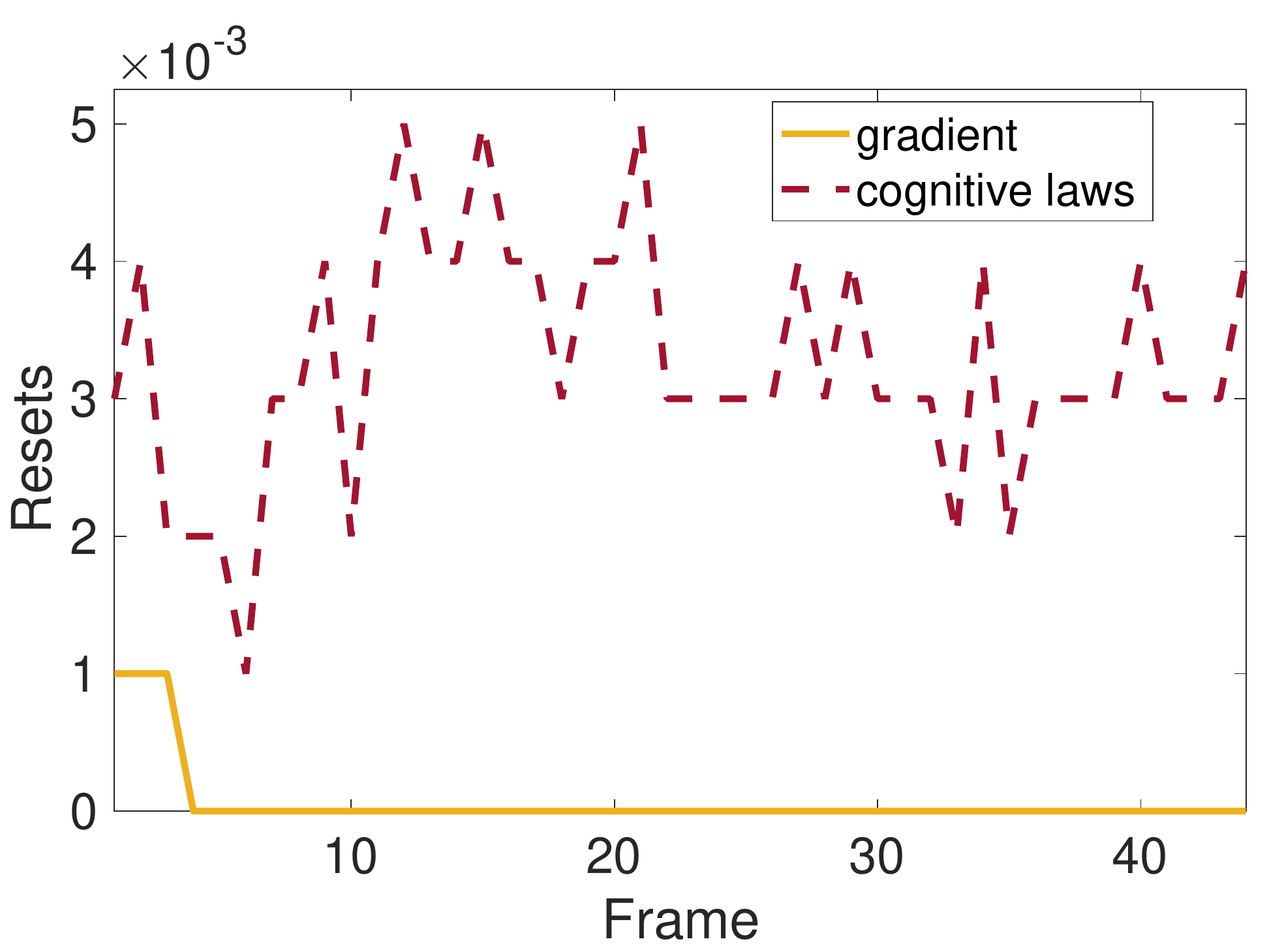|\hfil}
\vskip -3mm
\caption{We compare the cognitive laws with their special instance that resembles a stochastic gradient-based procedure ($n=10$ and filters of size $5\times5$).}
\label{grad}
\end{figure}

Finally, we experimented the setting of Eq. (\ref{reduction-gradient}), thus simulating an online gradient-descent with $\theta = 1000$. We generated an artificial video from the ``skater'' clip, by concatenating 3 instances of it, each of them using one of the R-G-B channels only. 
Fig.~\ref{grad} and Table \ref{mivw} (d) show that the gradient-like case leads to a smaller MI index and to an unstable evolution of it. We found that the MI is zero when reproducing the portions of video composed of shades of red or blue. This suggests that the system has focussed on features that are only about the greenish portion of the video, and that it was not able to capture information from the rest of the video, due to the large $\theta$. However, since there is only one derivative involved, the number of reset operation is almost zero.

\begin{table}[t]
  \caption{MI on the whole video, given the filters developed by
    the systems of Fig. \ref{smallk}. S=stability, R=reality, while the overline symbol indicates the lack of a property.}
    \vskip -1mm
\hbox to \hsize{\vbox{
\tabskip 3 pt plus 2 pt minus 2 pt
\hbox to .5\hsize{\hfil(a)\hfil}
\halign to.5\hsize{\hfil$#$\hfil&\hfil$#$\hfil&\hfil$#$\hfil&\hfil$#$\hfil&\hfil$#$\hfil\cr
\textup{FPS}&\bar S\bar R&\bar S R&R \bar R&SR\cr
\noalign{\smallskip\hrule\medskip}
25& 0.58 & 0.30 & 0.34 & \textbf{0.63}  \cr
10& \textbf{0.52} & 0.32 & 1.18\cdot10^{-5} & 0.50 \cr
100& 0.59 & \textbf{0.65} & 0.37 & 0.61 \cr
}\null\vfill}\hfill
\vbox{
\hbox to .45\hsize{\hfil(b)\hfil}
\tabskip 6 pt plus 2 pt minus 2 pt
\halign to.45\hsize{\hfil
\textsl{#}&\hfil$#$\hfil&\hfil$#$\hfil&\hfil$#$\hfil\cr
\textup{Video}\hfil&\textup{$n$=5, $5\times 5$}&\textup{$n$=11, $11\times 11$}\cr
\noalign{\smallskip\hrule\medskip}
Car & 0.39 & 0.28 \cr
Matrix & 0.62 & 0.46 \cr
Skater & 0.56 & 0.29 \cr
}\null\vfill}
\hfil}

\hbox to \hsize{\vbox{
\tabskip 3 pt plus 2 pt minus 2 pt
\hbox to .5\hsize{\hfil(c)\hfil}
\halign to.5\hsize{\hfil\textsl{#}&\hfil$#$\hfil\cr
\textup{Blurring}\hfil&\textup{$n$=11, $11\times 11$}\cr
\noalign{\smallskip\hrule\medskip}
Slow & 0.29  \cr
Fast & \textbf{0.41} \cr
Fastest (None) & 0.26  \cr
}\null\vfill}\hfill
\vbox{
\hbox to .45\hsize{\hfil(d)\hfil}
\tabskip 6 pt plus 2 pt minus 2 pt
\halign to.45\hsize{\hfil
\textsl{#}&\hfil$#$\hfil&\hfil$#$\hfil&\hfil$#$\hfil\cr
\textup{Video}\hfil&\hfil\textup{$n$=10, $5\times 5$}\cr
\noalign{\smallskip\hrule\medskip}
Gradient & 0.48  \cr
Cognitive Laws & \textbf{0.54} \cr
&\cr
}\null\vfill}
\hfil}
\label{mivw}
\vskip 0mm
\end{table}

\def\|#1|{\includegraphics{#1}}



\section{Conclusions}
In this paper we have proven that the minimization of the cognitive action 
yields causal equations, referred to as the  Cognitive Action Laws (CAL).
Depending on the choice of the potential they model different
tasks, while the choice of the kinetic energy characterizes the system dynamics,
and nicely extends the classic gradient flow and, consequently, the stochastic gradient descent. 
The main results that arises from the analysis of causal optimization is that the 
learning process must be driven by the gradual presentation of
the input that, in particular, when turned to zero, yields the reset of the dynamics. 
This sheds light on the video blurring process in newborns, which
stimulated the preliminary validation with the task of visual feature extraction presented in this paper. 
While the experiments are currently exhibited on simple video, it is worth mentioning that
the proposed approach is naturally suited for carrying out learning and inference by
an agent which is ``living in its own environment'' on a continuous video stream, 
where there is no difference between learning and test set. The enforcing
of the reset in the system dynamics might be very well suited for processing the 
video on the basis of the focus of attention as proposed in~\cite{DBLP:conf/nips/ZancaG17}.
When considering 
the very nature of this unsupervised learning scheme integrated with motion 
invariance~\cite{DBLP:journals/corr/abs-Betti2018}, the proposed theory 
opens the doors to a new approach to learning on video streams with no supervision.
This is especially interesting when considering deep convolutional architectures, 
that can be modeled by cognitive action laws by an opportune choice of the potential in the 
cognitive action.


\appendices

\section{Proofs of theorems}
\noindent
{\bf Proof of Theorem~\ref{GlobalMinimaRes}}:
The proof can be readily adapted from Lemma~2.1 of~\cite{Stefanelli2013}.
Because
of the terms proportional to $\mu$, $\nu$ and $k$ in
Eq.~(\ref{cognitive-action-2}) any minimizing sequence of this functional
is bounded in $H^2((0,T), \R^n)$; hence it is compact in $L^2((0,T), \R^n)$.
This motivates us to choose the following notion of convergence in
$L^2((0,T),\R^n)$:
\begin{equation}
q_{\kappa}\to q\,\hbox{(strongly)},\quad
\dot q_{\kappa}\to \dot q\,\hbox{(strongly)},\quad
\ddot q_{\kappa}\rightharpoonup \ddot q\,\hbox{(weakly)}.
\label{notion-of-convergence}
\end{equation}
These arguments accounts for the coercivity of the functional (compactness
of sub-levels).
For the lower semicontinuity the only adjustment we have to do is to show that
the term $\gamma\int_0^T e^{\theta t} \dot q\cdot \ddot q\, dt$ is
lower semicontinuous with respect to Eq.~(\ref{notion-of-convergence}).
This, however, can be  deduced from
a well known result that states that in a separable Hilbert space $H$ with
$v_n\rightharpoonup v$ and $w_n\to w$ and $\langle v_n\rangle$ limited,
the scalar product $(v_n, w_n)\to(v,w)$ since:
\[\begin{split}
    |(v_n,w_n)-(v,w)|\le &|(v_n,w_n)-(v_n,w)|\\
    +&|(v_n,w)-(v,w)|\le M\Vert w_n-w\Vert<\epsilon.
\end{split}\]
This being done we can conclude as in~\cite{Stefanelli2013}.

\medskip
\begin{remark}
  In the proofs of Theorems~\ref{vanishing-derivatives} and~\ref{MemoryDNS}
  we will use the fact that  the roots of the
  characteristic polynomial $\chi(x)$ of Eq.~(\ref{dyn-in-B})
  can always be chosen to be distinct, real, and to be
  more negative than any constant $-L$
  with $L>0$ and always bigger than $-L-\tau$ ($\tau>0$). The property of
  having real and distinct roots can be achieved by imposing condition~1
  of Lemma~\ref{reality-lemma}, while the other two more conditions
  coincide with the  Routh-Hurwitz criterion applied to $\chi(x-L)$ and
  to $\chi(-x-(L+\tau))$. We have checked for the existence of
  a solution to all this conditions using the licensed software
  Wolfram Mathematica~\cite{Mathematica}.
\end{remark}

\smallskip\noindent
{\bf Proof of Theorem~\ref{vanishing-derivatives}}:
  The dynamic in each $B_i$ is given by Eq.~(\ref{dyn-in-B}); since one of the
  roots of the characteristic polynomial is $0$, the solution in $B_i$ is thus
  $q(t)=c_1+c_2 e^{\lambda_{2}(t-t_{2i})} +c_3 e^{\lambda_{3}(t-t_{2i})}
  +c_4 e^{\lambda_{4}(t-t_{2i})}$, where $\lambda_i$ with $i=2,3,4$ are the
  remaining roots. In order to show the theorem it is sufficinet to show that
  $|q^{(k)}(t_{2i+1})|$ can be made arbitrary small for $k=1,2,3$. Indeed
  since it is possible to show that the coefficients in Eq.~\ref{dyn-in-B}
  can be choosen in such a way to have $\lambda_i<0$ and with
  $|\lambda_i|>L$ for any $L$ positive and $i=2,3,4$, the magnitude of the 
  derivatives
  \[\begin{split}
    |q^{(k)}(t_{2i+1})|=&|c_2\lambda_2^k e^{\lambda_{2}(t_{2i+1}-t_{2i})}
    +c_3\lambda_3^k e^{\lambda_{3}(t_{2i+1}-t_{2i})}\\
    &+c_4\lambda_4^k e^{\lambda_{4}(t_{2i+1}-t_{2i})}|,\quad k=1,2,3,\\
  \end{split}\]
is exponentially suppressed.

\smallskip\noindent
{\bf Proof of Theorem~\ref{MemoryDNS}}:
The proof arises when imposing the continuity of the derivatives at the time
$t=t_i$. Suppose that we have already solved the problem in $A_i$ and
therefore we know the values
$q(t_i),\dot{q}(t_i),\ddot{q}(t_i)$ and $q^{(3)}(t_i)$.
Since $\lambda_{1}=0$, the evolution in $B_i$ is given by
$q(t)=c_1+c_2 e^{\lambda_{2}(t-t_i)} +c_3 e^{\lambda_{3}(t-t_i)}
+c_4 e^{\lambda_{4}(t-t_i)}$.
Hence for $t=t_i$ we must have
$c_1+c_2+c_3+c_4=q(t_i)$ and
\[ 
	\begin{pmatrix} 
		\lambda_{2} & \lambda_{3} & \lambda_{4} \\ 
		\lambda_{2}^{2} & \lambda_{3}^{2} & \lambda_{4}^{2}  \\ 
		\lambda_{2}^{3} & \lambda_{3}^{3} & \lambda_{4}^{3} 
	\end{pmatrix}
	\begin{pmatrix}
		c_2 \\ c_3 \\ c_4
	\end{pmatrix}
	=
	\begin{pmatrix}
		\dot{q}(t_i) \\
		\ddot{q}(t_i)\\
		q^{(3)}(t_i)
	\end{pmatrix}.
\]
Let  
$V(\lambda_2,\lambda_3,\lambda_4)\equiv V_3$ be the above Vandermonde matrix.
Since it is always possible to choose the coefficients of Eq.~(\ref{dyn-in-B})
so that $\lambda_i<0$ and $-L-\tau<\lambda_i<-L$ for all $L>1$
and $\tau$ positive,
and since
$|q(t_{i+1}) -q(t_i)| = |c_2 (e^{\lambda_2 t_{i+1}}-1)
+c_3 (e^{\lambda_3 t_{i+1}}-1)+c_4 (e^{\lambda_4 t_{i+1}}-1)|$,
if we let $\rho=\max_{k} |\lambda_k|$ so that
$1/(1+\tau)<-\lambda_i/\rho<1$ we have
\[
\begin{split}
|q(t_{i+1}) -q(t_i)|&\le |c_2|+|c_3|+|c_4|\\
	&\le\sum_{j=1}^3\sum_{k=1}^3\rho^{-k}|\Lambda_{kj}|\cdot|q^{(k)}(t_i)|
        \\
        &\le 3 \frac{(1+\rho^2+\rho^4)}{\rho^6} C
        \max_k\cdot|q^{(k)}(t_i)|\\
        &\le 9 \rho^{-2} C\max_k
        |q^{(k)}(t_i)|,
        \end{split}
      \]
      where
      $\Lambda=(V(\lambda_2/\rho,\lambda_3/\rho,\lambda_4/\rho))^{-1}$,
      and $C>0$ is such that $|\Lambda_{kj}|\le C$ for
      all $k$ and $j=1,2,3$.

      

\bibliographystyle{IEEEtran}
\bibliography{nn,References}

\end{document}